\pdfoutput=1
\documentclass{article}

\usepackage{microtype}
\usepackage{placeins}
\usepackage{graphicx}
\graphicspath{{./}{../}}
\usepackage{booktabs}
\usepackage{amsmath}
\usepackage{float}
\usepackage{caption}

\usepackage{xcolor} 
\usepackage{tabularx}
\usepackage{wrapfig}

\usepackage[main,final]{neurips_2026}

\usepackage[colorlinks=true, linkcolor=blue, citecolor=blue, urlcolor=blue]{hyperref}

\newcommand{\num}[1]{#1}

\newcommand{\totalCEG}{3,360\times}              
\newcommand{\lstmKaplanCEG}{352\times}           
\newcommand{\chinchillaCEG}{3.7\times}           
\newcommand{\scaleInvCEG}{2.6\times}             
\newcommand{\scaleDepShare}{88\%}                
\newcommand{\scaleDepAbs}{1,309\times}           
\newcommand{\annualCEG}{2.1\times}              
\newcommand{\transitionShare}{72\%}              
\newcommand{\smallScaleLSTMtoTM}{7.9\times}     
\newcommand{\smallScaleTotal}{75.8\times}        
\newcommand{\apparentGap}{290\times}             
\newcommand{\gapToHo}{6.5\times}                
\newcommand{\transformerSeventeenCEG}{77\times}  
\newcommand{\modernTMExponent}{0.093}            
\newcommand{\interactionPredictedCEG}{21.7\times} 

\title{On the Origin of Algorithmic Progress in AI:\\
{\large Evidence from Language Model Pre-Training}}

\author{
  Hans Gundlach$^{\dagger,*}$ \\
  \texttt{hansgund@mit.edu}
  \And
  Alex Fogelson$^{\dagger}$ \\
  \texttt{fogelson@mit.edu}
  \And
  Jayson Lynch$^{\dagger}$ \\
  \texttt{jaysonl@mit.edu}
  \AND
  Ana Tri\v{s}ovi\'{c}$^{\dagger}$ \\
  \texttt{ana\_tris@mit.edu}
  \And
  Jonathan Rosenfeld$^{\dagger}$ \\
  \texttt{jonsr@csail.mit.edu}
  \And
  Anmol Sandhu$^{\ddagger}$ \\
  \texttt{asandhu@alumni.olin.edu}
  \AND
  Neil Thompson$^{\dagger,*}$ \\
  \texttt{neil\_t@mit.edu}
  \\ 
  \\
    $^{\dagger}$MIT FutureTech, CSAIL \quad
  $^{\ddagger}$Olin College \quad \\
  $^{*}$Corresponding authors
}

\begin{document}

\maketitle

\begin{abstract}
Algorithms have been estimated to increase AI pretraining FLOP efficiency by a factor of $22,000$ between 2012 and 2023 \citep{ho2024algorithmic}. Running ablation experiments on key innovations from this time period, we are able to account for less than $10\times$ of these gains at small compute scales. Surveying the broader literature, we estimate that additional innovations not included in our ablations also account for less than $10\times$, yielding less than $100\times$ efficiency gains total in low-compute regimes. This leads us to conduct scaling experiments, which provide evidence that the efficiency gap can be explained by algorithms with scale-dependent efficiency improvements. In particular, we conduct scaling experiments between LSTMs and Transformers, finding evidence for exponent differences in their compute-optimal scaling law while finding little scaling difference for many other innovations. These experiments demonstrate that -- contrary to standard assumptions -- an algorithm's efficiency gains are tied to compute scale. Using experimental scaling extrapolations and literature estimates, we can account for the majority of efficiency gains over the same time period. Our results indicate that algorithmic progress at smaller training compute levels has been far slower than previously assumed, and that measures of algorithmic efficiency are strongly reference-dependent.
\end{abstract}

\section{Introduction}
\label{sec:introduction}

Progress in AI over the past decade has been driven by two tightly coupled forces: rapidly increasing compute budgets and a succession of algorithmic efficiency innovations in architectures, optimization, and other training practices. While compute growth is relatively straightforward to measure, we still lack a clear accounting of algorithmic progress—what concrete changes drive efficiency gains, how large those gains are, and whether they persist across compute scales.

Recent work by \citet{ho2024algorithmic} has started to address this question. By analyzing hundreds of historical language models, they estimate that in the last 10 years, algorithmic progress has contributed more than 4 orders of magnitude in compute efficiency, while raw compute scaling has increased 7 orders of magnitude based on analyzing historic AI literature. Specifically, models have gotten roughly $22,000\times$ more efficient across the sum total of innovations in algorithms from 2012-2023, supposedly allowing the same performance level with drastically fewer FLOPs.

However, a precise decomposition of this progress remains largely unknown, and many questions about the origins of algorithmic advancement remain underexplored. For instance, how do algorithmic advances interact? Are algorithmic advances driven by a series of small improvements, or a few large ones? Do algorithmic improvements follow smooth, Moore's law-like trends, or are they driven by punctuated equilibria? Answering these questions has important implications for the future of AI, as it could enhance the predictability of future algorithmic gains, guide innovation efforts towards specific areas of improvement, and inform our expected returns to compute scaling.

In order to answer these questions, we take three complementary approaches: we conduct ablation experiments on important algorithmic advances in language models; we conduct scaling experiments to measure differences in optimal scaling across architectures; and we perform theoretical analysis of transitions in data and parameter scalings. Our experiments focus primarily on architectural algorithmic modifications, but we try to address other methods to improve efficiency, like tokenizers, using literature estimates. Our analysis points to three core conclusions:
\begin{enumerate}
    \item \textbf{We find most algorithmic innovations we experimentally evaluate have small, scale-invariant efficiency improvements} with less than \num{$10\times$} compute efficiency gain overall, and representing roughly \num{$10\%$} of total improvements at 2023 compute levels. This suggests that scale-invariant algorithmic progress contributes only a minor share of overall efficiency improvements (Section~\ref{sec:scale_invariant}).

    \item \textbf{We identify two scale-dependent sources of efficiency gains: the LSTM-to-Transformer transition and Kaplan-to-Chinchilla rebalancing.} Under our extrapolation assumptions, these account for \num{\(\scaleDepShare\)} of the reconstructed log-efficiency gains at 2023 compute levels. Our results suggest that gains estimated at large compute scales can substantially overstate those available at smaller scales (Section~\ref{sec:scale_dependent}).

    \item \textbf{We show that in the presence of scale-dependent innovations, not only do efficiency gains require continued compute investment, but the rate of algorithmic progress strongly depends on your choice of reference algorithm.} In other words, the rate of progress in successive models can appear exponential relative to one baseline algorithm, yet be zero relative to another (Section~\ref{sec:implications}).
\end{enumerate}

Collectively, these findings suggest that gains in algorithmic progress may be inherently scale-dependent, requiring ever-increasing compute to realize their benefits, and implying that algorithmic progress has benefited model builders with greater compute resources more than smaller players in AI development.

\subsection{Previous Work}
A growing literature measures the contribution of algorithmic advances to ML efficiency. In computer vision, \citet{hernandez2020measuring} found that the training FLOPs required to reach AlexNet-level performance fell by $44\times$ from 2012--2019. In language modeling, \citet{ho2024algorithmic} estimate roughly $22{,}000\times$ algorithmic efficiency gains from 2012--2023, though this magnitude has been questioned: \citet{whitfill2025note} argues that correcting for observational bias (notably, correlations between compute scale and measured efficiency) may reduce estimated progress by about an order of magnitude. Experimental decompositions of historical gains remain limited; \citet{sanderson2025rethinking} report modest improvements (e.g., $\sim 1.7\times$) from specific components such as rotary embeddings and layer normalization.

A central open question is whether algorithmic gains are scale-dependent. Several studies report little change in scaling exponents across architectures or optimizers \citep{hestness2017deep,bansal2022data}, and theory work often emphasizes data properties as the primary driver of scaling behavior and scale-dependent change \citep{rosenfeld2021scaling,michaud2023quantization,Bahri_2024}. Conversely, other empirical results suggest architecture can affect scaling: \citet{droppo2021scaling} find steeper scaling for Transformers than LSTMs in acoustic modeling, and \citet{sanderson2025rethinking} theorize that major transitions (e.g., LSTM$\rightarrow$Transformer) may have scale-dependent effects. Together with these studies, our work builds on the compute-equivalent gains framework of \citet{davidson2023aicapabilitiessignificantlyimproved} and \citet{ho2024algorithmic}, extending the comparison from a single multiplier to a function of compute. We test the scale-dependent transition hypothesized by \citet{sanderson2025rethinking} directly in language modeling, combining controlled component ablations with scaling comparisons among LSTMs, historical Transformers, and modern Transformers. We also do many ablations test and show that combined improvements can yield smaller gains than multiplying their individual effects would predict, and that measured rates of progress depend on the reference algorithm.

\section{Compute-Equivalent Gains with Scale Dependence}\label{sec:expanding_gains_framework}

To compare algorithmic choices in training efficiency, we use the compute-equivalent gains (CEG) framework
\citep{davidson2023aicapabilitiessignificantlyimproved,ho2024algorithmic,sanderson2025rethinking}, and extend it to allow scale-dependent improvements. For two training algorithms $A$ and $A'$ (where ``algorithm'' includes architectural changes such as LSTM-to-Transformer, as well as training-recipe and component changes such as activation functions, normalization, positional encodings, learning-rate schedules, optimizers, and data–parameter allocation, etc.), define a \emph{CEG function} $f(C)$ such that training with $A'$ using compute $C/f(C)$ matches the performance of training with $A$ using compute $C$. A \emph{CEG multiplier} is then the point evaluation $f(C_0)$ for a fixed reference compute $C_0$ (equivalently, at a fixed performance level). This distinction matters in our setting because scale-dependent innovations can make CEG multipliers appear to grow over time even when the underlying comparison between algorithms is not constant across compute. Formal definitions and edge cases are given in Appendix~\ref{app:ceg_formal}.

\section{Scale-Invariant Algorithms}
\label{sec:scale_invariant}

\begin{wrapfigure}{R}{0.44\linewidth}
    \vspace{-1em}
    \centering
    \includegraphics[width=\linewidth]{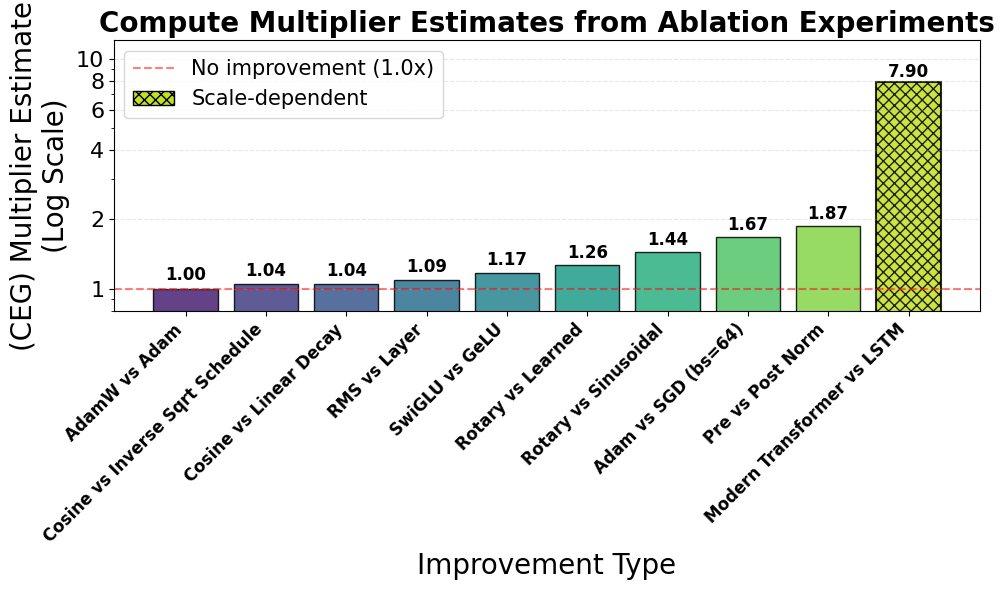}
    \vspace{-0.75em}
    \caption{Compute-equivalent gain (CEG) multipliers from ablations on a \textit{small} Transformer. Most component-level changes yield modest efficiency gains. Later experiments show only one ablation is scale-dependent.}
    \label{fig:bar_improvements}
    \vspace{-1em}
\end{wrapfigure}
We estimate the compute-equivalent gains (CEGs) of several widely used training choices via controlled ablation experiments. Concretely, we vary one component at a time (activation functions, positional encodings, normalization, learning-rate schedules, and optimizers) and measure the compute required to reach a fixed validation threshold, yielding CEG multipliers (Section~\ref{sec:expanding_gains_framework}). We also test whether improvements compose multiplicatively by ablating multiple components jointly (Figure~\ref{fig:interaction}; Appendix~\ref{app:ablation_details}).

Across the compute regime of our experiments, these changes appear largely scale-invariant: in Section~\ref{sec:scale_dependent} we find near-identical scaling exponents between our baseline Transformer and a Transformer with all post-2017 components ablated (Figure~\ref{fig:scaling_graphs}). Overall, the cumulative scale-invariant gains we measure are modest (under $10\times$ in total), with most individual changes contributing less than $2\times$. This motivates our focus in Section~\ref{sec:scale_dependent} on scale-dependent innovations.

\subsection{Ablation Experiments}
\label{sec:ablations}

We run ablations on a small Transformer baseline and compute CEG multipliers at a common validation threshold; full experimental details and per-component results are deferred to Appendix~\ref{app:ablation_details}. Figure~\ref{fig:bar_improvements} summarizes the measured multipliers. In brief, we find small improvements from activation functions and positional encodings, negligible differences across common learning-rate schedules, and larger (but still $<2\times$) gains from normalization and optimizer choice. We further analyze optimizer scaling separately in Appendix~\ref{sec:sgd_scaling}.

\begin{wrapfigure}{r}{0.49\linewidth}
    \vspace{-1.5em}
    \centering
    \includegraphics[trim=0 -23.04bp 0 0,height=0.55\linewidth]{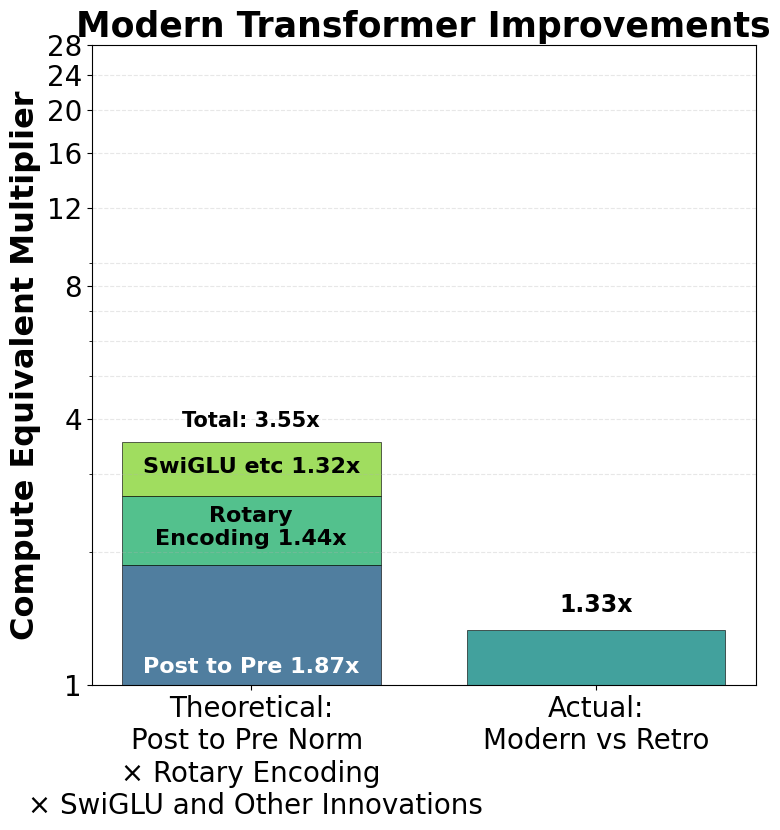}\hfill
    \includegraphics[height=0.55\linewidth]{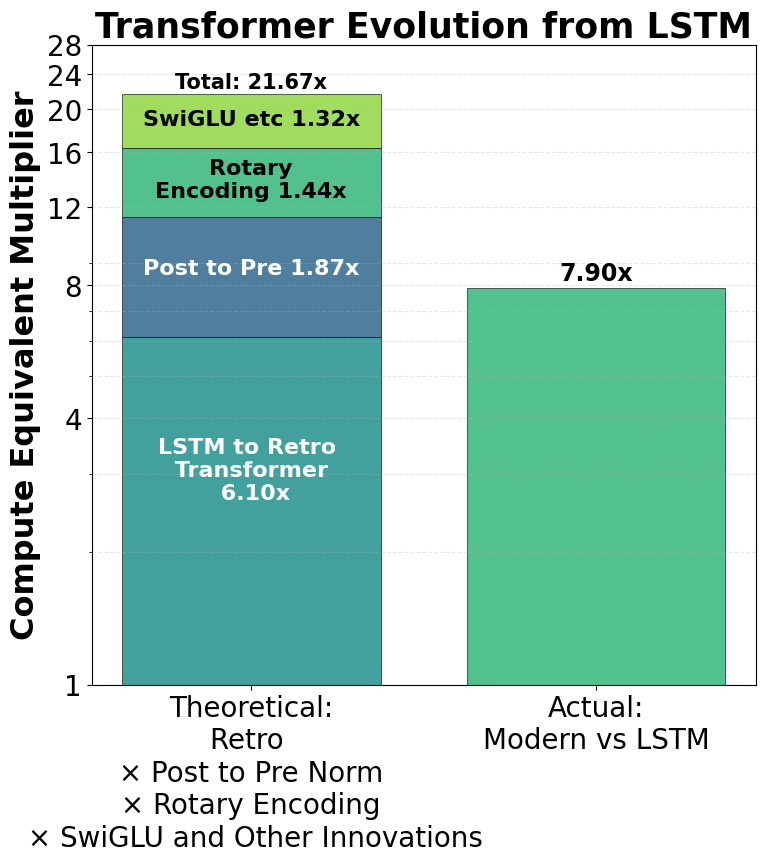}
    \vspace{-0.5em}
    \caption{Interaction between algorithms: measured combined gains of post-Transformer changes (left) and of all changes including the LSTM-to-Transformer transition (right) vs.\ the product of individual ablation estimates. Combined gains are sub-multiplicative (Appendix~\ref{sec:interaction_algs}).}
    \label{fig:interaction}
    \label{fig:modern_transformer_interaction}
    \label{fig:lstm_modern_interaction}
    \vspace{-2.25em}
\end{wrapfigure}

\paragraph{Interactions Between Improvements.}
 Combining the studied ablation improvements yields a compute-equivalent gain of \num{\(\smallScaleLSTMtoTM\)} relative to our LSTM baseline, compared with \num{\(\interactionPredictedCEG\)} predicted by multiplying individual gains. Thus, multiplying ablation-based estimates can overstate combined improvements (Figure~\ref{fig:interaction}). We discuss these interaction effects and their implications for theories of algorithmic progress in Appendix~\ref{sec:interaction_algs} and adjust for this in Figure~\ref{fig:stackplot}.

\paragraph{Literature-Based Components}
Some important contributors to efficiency are difficult to evaluate cleanly in our setup, namely mixture-of-experts (MoE) architectures, tokenizers, and innovations in data selection. When reporting efficiency gains over time, we judge literature estimates for tokenizers ($1.68\times$) and data ($\leq3\times$) to be highly uncertain, and therefore only include MoE ($2\times$) in our stylizations and aggregations. Further discussion on literature estimates can be found in Section~\ref{sec:data} and Appendix~\ref{app:literature_estimates}.

\paragraph{Implication: Small, Skewed Scale-Invariant Gains}
\label{sec:small_small}
At small scales, the aggregate effect of the scale-invariant components we measure is limited, and the distribution is skewed: a handful of changes account for most of the gains while many contribute little. This motivates the scaling experiments in Section~\ref{sec:scale_dependent}, where we test whether the largest historical gains arise from scale-dependent innovations rather than the accumulation of many scale-invariant tweaks.

\section{Scale-Dependent Algorithms}
\label{sec:scale_dependent}

Are algorithmic gains larger at larger compute scales? To answer this question, we conduct scaling experiments across architectures, optimizers, and the innovations from Section~\ref{sec:ablations} to get a better sense of how their efficiency gains scale with compute. In addition, we exhibit a mathematical analysis of the Kaplan to Chinchilla scaling shift, which gives a particularly interesting example of a scale-dependent algorithmic change that is not an exponent shift.

\paragraph{Experimental Setup for Scaling}
\label{sec:experimental_setup}

We compare three training algorithms: a \textbf{Modern Transformer}, a \textbf{Retro Transformer} that reverts key post-2017 components (Appendix~\ref{sec:interaction_algs}), and a \textbf{pre-Transformer LSTM baseline}. To separate \emph{architecture} from \emph{scaling policy}, we also define a \textbf{Kaplan Transformer} as the Retro Transformer evaluated under Kaplan-style parameter--data allocation via an analytical adjustment (Appendix~\ref{app:scaling_models_details}). Table~\ref{tab:model_orientation} summarizes the configurations; full architecture and hyperparameter details are deferred to Appendix~\ref{app:scaling_models_details}.

\begin{wraptable}{r}{0.48\linewidth}
\vspace{-1em}
\centering
\small
\setlength{\tabcolsep}{3pt}
\begin{tabular}{@{}lcc@{}}
\toprule
\textbf{Algorithm} & \textbf{Innov.} & \textbf{Scaling} \\
\midrule
Modern Transformer & Included & Optimal \\
Retro Transformer  & Excluded & Optimal \\
Kaplan Transformer & Excluded & Kaplan \\
LSTM baseline      & N/A      & Optimal \\
\bottomrule
\end{tabular}
\vspace{-0.5em}
\caption{Model configurations used in scaling experiments.}
\label{tab:model_orientation}
\vspace{-1em}
\end{wraptable}

\paragraph{Transformer baselines.}
Our Modern Transformer is a small, vanilla decoder-only Transformer with standard contemporary choices (e.g., rotary position encodings, RMSNorm prenorm, and SwiGLU), using a fixed width-to-depth aspect ratio appropriate for the small-model regime. The \textbf{Retro Transformer} matches the Modern Transformer except that we revert four components (activation, normalization placement/type, positional encoding, and learning-rate schedule) to approximate an early Transformer-style recipe. We keep these definitions fixed across scales and tune learning rates uniformly across model families; exact architectural choices and all hyperparameters are listed in Appendix~\ref{app:scaling_models_details}.

\paragraph{LSTM baseline}
We use a vanilla LSTM intended to reflect strong pre-Transformer practice (following \citet{melis2017state}) and exclude later variants (e.g., GRUs, Mogrifier, xLSTM). We use a \textbf{single LSTM layer} because (i) \citet{melis2017state} finds 1-layer LSTMs are competitive at $\sim$10M parameters, and (ii) in our scaling runs, 1-layer is consistently better than 2-layer at both the smallest and largest sizes we test and has sharper scaling exponent (see Appendix~\ref{app:2_layer_baseline} for experiments). 1-layer also aligns with the guidance in \citet{droppo2021scaling}. We adopt standard LSTM initialization and regularization conventions from \citet{melis2017state} (Table~\ref{tab:lstm_hyperparams}). Because our training regime uses large-scale web data where overfitting is minimal, we set LSTM dropout to zero (analogous to common Transformer practice in this setting \citep{hoffmann2022trainingcomputeoptimallargelanguage}); this materially improves validation loss in our compute-optimal runs (Appendix~\ref{app:scaling_models_details}). See Appendices~\ref{appendix:model_architecture} and~\ref{app:scaling_models_details} for further justification of our baseline.

\paragraph{Scale range and tuning.}
We scale each family by varying hidden size over a fixed grid (Appendix~\ref{appendix:model_architecture}), yielding Transformers spanning $\sim$1.6M--29M parameters and LSTMs spanning $\sim$1.6M--20.5M parameters. For each family, we run matched learning-rate tuning and fit a smooth function from sizes to learning rates across scales (Appendix~\ref{sec:learning_rate_tune}).

We scale both the Modern Transformer and the Retro Transformer; the latter ablates the post-2017 algorithmic changes described above. For each setup, we follow conventional scaling practices, keeping depth-to-width ratio constant.\footnote{See Appendix~\ref{appendix:model_architecture} and the following discussion for more details on our experimental setup.
}

\begin{figure*}[t]
    \centering
    \includegraphics[width=0.8\textwidth]{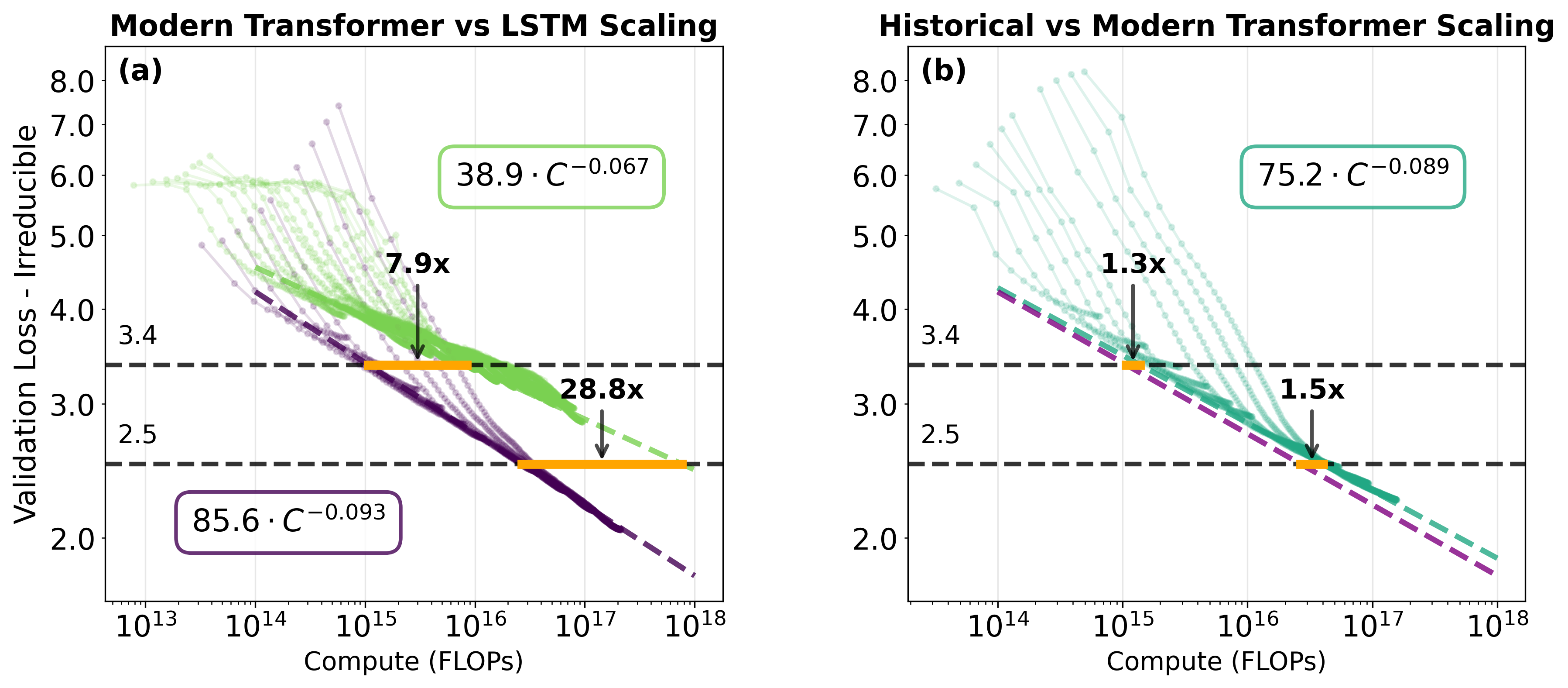}
    \caption{Figure (a) depicts the scaling difference between a Modern Transformer in purple and a standard LSTM in green. Figure (b) depicts the scaling difference between a Modern Transformer in purple and a Retro Transformer in blue, where all post-2017 innovations are ablated. LSTMs seem to have significantly different scaling exponents, while post-2017 transformers have minimal effect on scaling. All graphs depict training curves for models with hidden dimensions between 32 and 256 (384 for LSTMs) with all other hyperparameters scaled proportionately.}
    \label{fig:scaling_graphs}
\end{figure*}

\subsection{Transformers are a Scale-Dependent Improvement to LSTMs}

We plot the difference in scaling between LSTMs and Modern Transformers in Figure~\ref{fig:scaling_graphs}A and between Modern and Retro Transformers in Figure~\ref{fig:scaling_graphs}B. For each, we subtract the irreducible loss component (1.9, see Appendix~\ref{sec:irreducible_loss}) and fit a power law to all Pareto-optimal (frontier) points from $10^{16}$ FLOPs onward (fit statistics in Appendix~\ref{app:fit_statistics}).\footnote{Additional scaling experiments for Adam-to-SGD are described in Appendix~\ref{sec:sgd_scaling}.}

Our scaling graphs suggest that improvements in neural network architecture are not scale invariant and have increasing returns to scale. Interestingly, using an irreducible loss-adjusted threshold of 3.4, our LSTM model exhibits $7.9\times$ less compute-efficiency than our transformer model, while a threshold of 2.5 yields an efficiency gap of $29\times$ (Figure ~\ref{fig:scaling_graphs}). At 2017 compute levels, the efficiency gain is approximately \num{$\transformerSeventeenCEG$}, slightly larger than previous estimates from the literature at these scales. \citet{ho2024algorithmic} attributes a $60\times$ gain to this transition. \citet{sanderson2025rethinking} estimates that this transition had algorithmic FLOP efficiency gains of $20-50\times$ at scales of around $10^{19}$ FLOPs.

Meanwhile, the choice between Adam and SGD, as well as the difference between a Modern and a Retro Transformer, seems to be scale-invariant. This provides direct evidence that optimizers may exhibit constant CEG functions, and strong evidence that our ablated algorithms would not individually show scale-dependencies.

\subsection{Transition from Kaplan to Chinchilla Scaling Laws}

In our discussion so far, we have focused on model training choices and have paid relatively little attention to how models are scaled, assuming they are scaled optimally between dataset size and parameter count for a set of algorithms. However, the shift from so-called ``Kaplan" to ``Chinchilla" scaling practices are a particularly interesting example of scale-dependent algorithmic progress that is not an exponent shift.

\citet{kaplan2020scalinglawsneurallanguage} advocates for a smaller level of data scaling compared to \citet{hoffmann2022trainingcomputeoptimallargelanguage}. Subsequent literature found that Kaplan misestimated scaling law exponents due to small sizes, not accounting for embedding parameters, and using a constant warmup size \citep{porian2024resolving,pearce2024reconciling}. In contrast, \citet{hoffmann2022trainingcomputeoptimallargelanguage} accounted for all parameters and found that data and parameters should be scaled equally while scaling compute (so-called ``Chinchilla-scaling"). Our analysis provides a closed form for the CEG function between Kaplan and Chinchilla scaling similar to \citep{ho2024algorithmic}.

To formalize this switch from Kaplan to Chinchilla laws, we take Chinchilla-optimal scaling laws as ground truth, and assume model trainers allocate parameters and data according to misspecified Kaplan scaling laws (see Appendix~\ref{appendix:kaplan_chinchilla}). This form of algorithmic improvement has clear scale-dependent effects, though not a simple exponent change. Training efficiency initially converges between the two scaling laws (within the regime tested by Kaplan), but diverges at larger scales (Figure~\ref{fig:kaplan-v-chinch}). However, the Chinchilla training efficiency gains remain below \(10\times\) across much of the compute range used in contemporary machine learning. This estimate assumes model developers were using Kaplan and Chinchilla scaling historically. It is unclear for many models what scaling recipe they were using, so this does not provide causal evidence. However, this provides evidence that such a shift could be another important scale-dependent driver of algorithmic progress.

\begin{figure*}[t]
    \centering
    \includegraphics[width=0.9\textwidth]{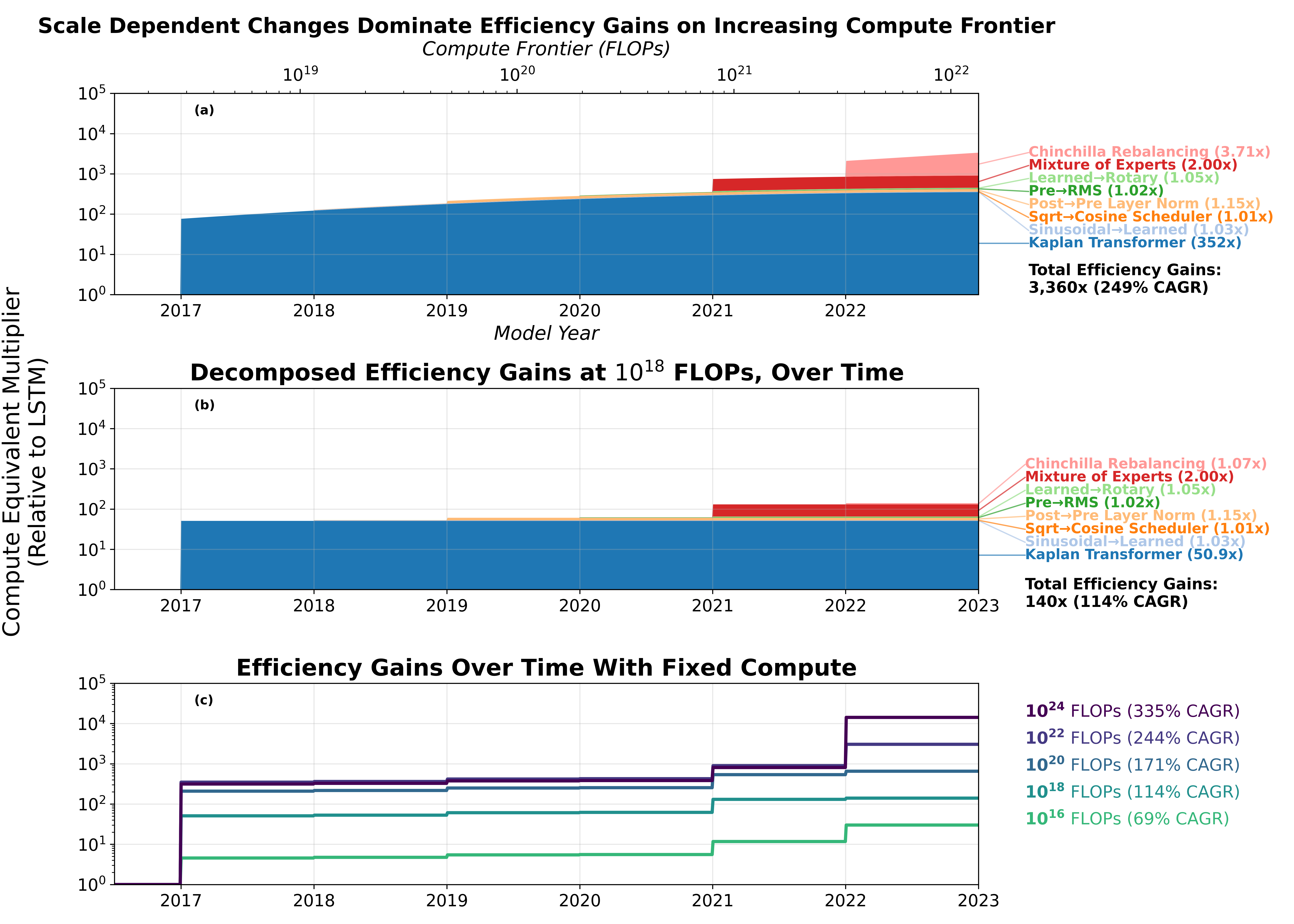}
    \caption{LSTM to Transformer transition has been responsible for most of the algorithmic progress out of the innovations we review. We adjust scale-invariant transitions to be log-proportional to our overall ablation results, given the interaction effects we demonstrate in Appendix~\ref{sec:interaction_algs}. }
    \label{fig:stackplot}
\end{figure*}

\section{Algorithmic Progress Depends Strongly on Compute, Reference Points}

Our theoretical and experimental results allow us to decompose algorithmic progress into individual innovations introduced over time. This leads to two main observations. First, many measured CEG gains may be a consequence of compute scaling rather than a multitude of algorithmic innovations. Second, when innovations are scale dependent, the growth rates of CEG multipliers depend entirely on the chosen reference model, meaning that progress may appear unbounded with respect to one algorithm while non-existent with respect to another.

\subsection{Compute (Not Time) May Explain Most Algorithmic Progress}

Existing estimates of algorithmic progress measure the rate of progress in algorithms as a function of time \citep{ho2024algorithmic}. However, our analysis opens the question as to whether this time-dependence is instead driven by steady increases in compute investment. Estimates show the compute budgets of notable models have exponentially increased at a rate of $4.1\times$ per year~\citep{epoch2024trainingcompute}. Thus as compute budgets increase exponentially, the rate of algorithmic progress may in fact be driven more by the regularity of compute scaling rather than the discovery of new innovations. 

We stack our empirical results, as well as literature estimates for mixture of experts and theoretical results, to extrapolate the expected CEG gains over varying compute scales (with respect to LSTMs). To compare our results to literature estimates (which track gains over time), we use the (aforementioned) exponential relationship between time and compute among notable models. We denote this trend as the ``compute frontier" (distinct from the steeper trajectory one obtains by considering only the top-performing model at each date). This allows us to calculate the cumulative CEG function over historically relevant time intervals.

When stacking our ablated, scale-invariant innovations, we scale down their total CEG multiplier according to our interaction experiments (Appendix~\ref{sec:interaction_algs}). When reporting the decomposition, the results remain log-proportional to the independent multipliers. Moreover, when reporting gains from Kaplan Transformers, we analytically adjust our Retro Transformer according to gains from the Kaplan-to-Chinchilla CEG function.

Importantly, between 2017 and 2023, we provide extrapolated evidence that the majority of algorithmic progress could be accounted for by two scale-dependent innovations: the switch from LSTM to Kaplan Transformers and the rebalancing to Chinchilla scaling. Of the total measured progress our extrapolations account for \num{$\totalCEG$} (relative to an LSTM); we find that \num{$\lstmKaplanCEG$} could be achieved through LSTMs to Kaplan Transformers, and  \num{$\chinchillaCEG$} could be accounted for by Chinchilla rebalancing. Together, these comprise \num{$\scaleDepShare$} of our extrapolated total log-efficiency gains, the vast majority of our measured progress.  However, these two innovations retain a majority share of the accounted log-gains across the tested $\pm25\%$ perturbations of the exponent gap. Although we try to measure the scaling exponents as accurately as possible, exact multipliers could be sensitive to the exponent in LSTM scaling (see here for the sensitivity discussion Appendix~\ref{sec:sensitivity_analysis})

\begin{wrapfigure}{r}{0.49\linewidth}
    \vspace{-1.1em}
    \centering
    \includegraphics[width=\linewidth]{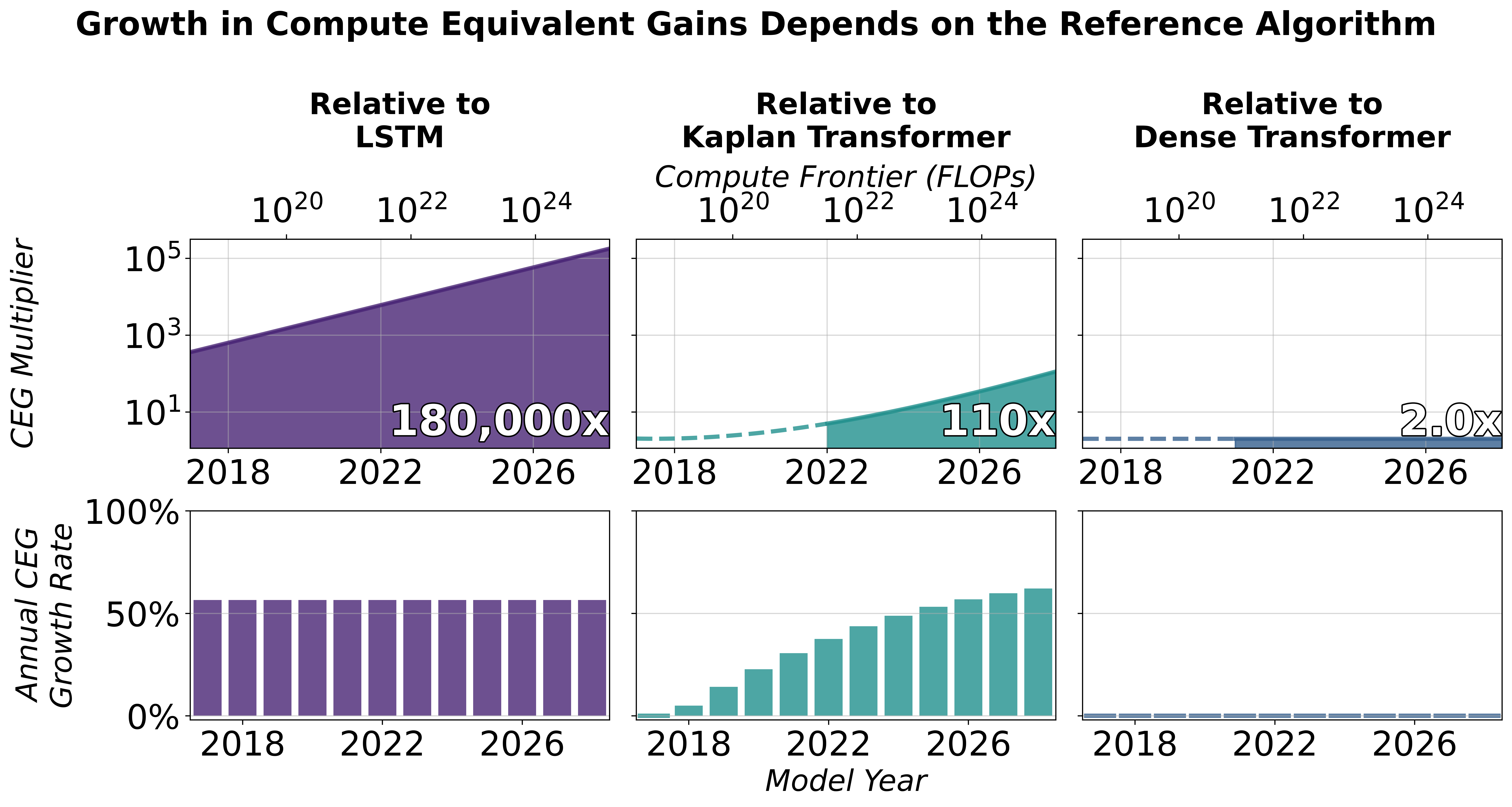}
    \vspace{-1.25em}
    \caption{Top: CEG gain of a MoE Transformer relative to LSTM, Kaplan Transformer, and dense Transformer. Bottom: corresponding annual CEG growth rates, extrapolated to 2028; they vary dramatically with the reference choice. Years are aligned with approximate historical model sizes.}
    \label{fig:reference_models}
    \vspace{-1.75em}
\end{wrapfigure}
Though our experiments do not claim to be exhaustive, we compare our findings with estimates from the literature. Namely, between 2012 and 2023, \citet{ho2024algorithmic} found a doubling time of $8$ months, or \num{$2.83\times$} per year, for a total efficiency gain of $22,000\times$. In contrast, the growth rate of our extrapolated CEG multiplier is approximately \num{$\annualCEG$} annually over the same time period, for a total of \num{$\totalCEG$}, of which \num{$\scaleDepAbs$} (\num{$\scaleDepShare$}) of log-efficiency gains are due to scale-dependent changes. This leaves a gap of \num{$\gapToHo$} from our estimates, which could be from data selection, the sensitivity of our results to exponent change, tokenizer advancements, or a long tail of innovations not captured in our analysis. We include a sensitivity analysis of our results in Appendix~\ref{sec:sensitivity_analysis}.

\subsection{CEG Multipliers Depend Strongly on Reference Algorithms}
\label{sec:implications}

Once we introduce scale-dependent algorithms, \textit{a deeper problem exists in how we measure algorithmic progress}. 

Consider a sequence of Modern Transformer models $M_1,M_2,..., M_t$, each enhanced with a mixture-of-experts architecture, and trained with exponentially increasing compute. If we want to measure the growth rate of CEG multipliers (i.e., the rate of algorithmic progress), we need to choose a reference algorithm, such as baseline LSTMs or dense transformers. Measuring the CEG multiplier with respect to LSTMs for each model $M_t$ yields an exponential in $t$, with a growth rate of about $63\%$ annually. However, measuring with respect to the dense Transformers yields a constant multiplier of \num{$2\times$} (see Appendix~\ref{app:moe} for literature estimates), and therefore a $0\%$ growth rate. Thus choosing one reference point yields an \textit{exponential growth} rate in algorithmic efficiency, while a different reference point yields \textit{zero growth}. Figure~\ref{fig:reference_models} visualizes this reference dependence, and also shows the apparent growth in CEG multiplier from the Kaplan-Chinchilla transition (see also Appendix~\ref{app:reference_dependence}).

Only once scale-dependent algorithms are considered, selecting different reference algorithms for computing CEG multipliers can determine whether growth appears exponential, non-existent, or somewhere in between. In this sense, algorithmic progress with scale-dependent innovations is strongly contextual: even if you are completely certain you will measure exponential growth rates in the future with respect to some \textit{fixed algorithm}, this offers no guarantee of any progress at all \textit{relative to current models}. Importantly, this \textit{is not} merely the claim that ``we cannot be assured algorithmic progress will continue." Rather it is a statement about the measurement framework: one can continue to \textit{measure} arbitrarily large rates of progress from some fixed point while realizing none of these gains in practice.

\section {Limitations and Further Work}

\paragraph{Impacts of Data Selection}
\label{sec:data}
We do not do any dataset ablation experiments. In general, it is hard to compare performance between datasets as perplexity is a dataset-relative benchmark. Further, frontier models have been training at ever larger dataset sizes which might necessitate using lower quality-data making extrapolation claims from small datasets hard. However, it is generally thought that datasets play an important role in determining neural scaling law exponents \citep{rosenfeld2021scaling, michaud2023quantization}. \citet{sorscher2022beyond} found that neural scaling laws can even be exponential rather than power laws under the right data pruning metric. Nevertheless, in practice, while a better dataset may improve power laws, the difference between most datasets on evaluation metrics is moderate. \citet{li2024datacomp} did a large-scale study comparing pretraining efficiency on new ($<$ 2024) and historical datasets. They claim a $10\times$ efficiency gain for their DCLM baseline dataset, but most historical web datasets are within a factor of $3\times$ FLOP efficiency of one another in their study at scales of $10^{20}$ to $10^{22}$ FLOPs. However, the impact of data and scale-dependent data change is an active area of research, and new datasets such as UltraFineWeb \citep{wang2025ultra} could have a larger impact \citep{patel2026pretraining}.

\paragraph{Our Experiments are Relatively Small and Non-Exhaustive}

Our experiments are conducted at small scales compared to more recent scaling studies (i.e., \citet{hoffmann2022trainingcomputeoptimallargelanguage}). However, our studies are done at slightly lower but similar scales to papers that originally explored many of these innovations, like \citet{melis2017state}.

Finally, there are many things we did not test experimentally. These include mixture of experts, tokenizers, improvements in data, and other innovations which may be underrepresented in the literature or less widely known. These are important contributors to algorithmic progress, but we are only able to give literature-based estimates. We also do not test innovations involved in the creation of LSTMs. However, many LSTM variations have little benefit \citep{greff2016lstm}, and we use a generic architecture that would have been available in 2012. Further, we did not run scaling interventions on all the algorithms in our study individually, though when we ablate all algorithmic improvements, we find very small exponent differences in scaling exponents. We conclude from this that individual innovations admit negligible exponent shifts alone. Our study stops at innovations in 2023. Since then, there have been improvements in scaling recipes (like scaling expert number for MoEs) as well as reinforcement learning, which might have further scale-dependent effects \citep{tillman2026scaledependent}. Recent innovations in agentic scaling and scaffolding could also have scale-dependent effects, but fall outside the scope of our study.

\paragraph{The Burden of Optimality}
Another key limitation is the fickle nature of hyperparameters. For instance, we found it nearly impossible to scale a transformer trained with SGD and a batch size over 128. We are more confident in our results because our performance and exponents align well with other studies like \citet{porian2024resolving} and \citet{melis2017state}. Notably, \citet{porian2024resolving} conducts Chinchilla optimal scaling and finds scaling exponents from $.092$ to $.106$. Using the same tokenizer but slightly different data than \citet{porian2024resolving}, we achieve a scaling exponent of $\modernTMExponent$. \citet {hoffmann2022trainingcomputeoptimallargelanguage} achieves a scaling exponent of $.154$ but uses both a different tokenizer and different data. 

However, our study leaves open the possibility of ``golden" hyperparameters that could significantly lower the gap between the algorithmic improvements we measure or even change the experimental scaling relationship. For instance, SGD was long thought to be inherently inferior to Adam for transformer training, but studies like \citet{sreckovic2025your} demonstrate a very small performance gap with the right hyperparameters. We found very little literature on scaling LSTMs. Therefore, we could find improved scaling for LSTMs if extensive hyperparameter sweeps were done at all scales. This becomes computationally infeasible with much larger models, as determining optimal scaling for transformers took many years of progress in the machine learning community. Finding such an optimal scaling for other architectures and settings might require similar effort. 

\paragraph{Beyond FLOP Efficiency: Reasoning and Other Dimensions of Progress}
Reasoning-optimized language models have recently emerged as a new pathway for improving capabilities, with developers increasingly optimizing models to ``think longer.'' \citet{epoch2025quantifyingthealgorithmicimprovementfromreasoningmodels} estimate that the shift from conventional to reasoning models is equivalent to a $10\times$ gain in pretraining compute, though these gains come with substantially higher inference costs. These gains along with newer developments in reinforcement learning, fall outside the scope of our study. 

More broadly, our focus on training FLOP efficiency necessarily omits many critical innovations. Several changes we test (e.g., RMSNorm and learning-rate schedules) show modest FLOP-efficiency gains but materially improve stability, convergence reliability, and trainability. Other advances reduce wall-clock time and energy without changing FLOP counts (e.g., Flash Attention), while capability-enabling methods such as instruction fine-tuning \citep{wei2021finetuned} and constitutional AI change model utility without improving pretraining efficiency. Finally, systems advances (e.g., pipeline and tensor parallelism) make large-scale training more feasible~\citep{barnett2025compute}. Taken together, these examples suggest that algorithmic progress is only partially captured by scalar FLOP-efficiency metrics, and is better understood along multiple axes including efficiency, stability, capability, and deployment cost.

\section{Discussion}
\label{sec:discussion}

Our experimental and theoretical analysis reveals that algorithmic progress in language models exhibits fundamentally different behavior across compute scales, with implications for understanding historical progress and future trajectories. While there is still much work to be done understanding data and more contemporary algorithms, we can analyze what our current results suggest.

\paragraph{Reconciling Small-Scale Measurements with Large-Scale Progress}

At performance thresholds comparable to Transformers trained with $10^{15}$ FLOPs (i.e., NLL = 5.3), we measure total algorithmic efficiency gains of approximately \num{$\smallScaleLSTMtoTM$} relative to baseline LSTMs when accounting for all tested innovations. Combined with literature estimates -- MoE estimated at ($\sim 2\times$), improved datasets at ($\sim 3\times$), and tokenizers at ($\sim 1.6\times$) -- this totals \num{$\smallScaleTotal$}, well under $100\times$. This contrasts sharply with \citet{ho2024algorithmic}'s estimate of $22,000$ efficiency gains over the last decade, leaving an apparent gap of \num{$\apparentGap$} or more.

However, this discrepancy could be resolved by scale-dependent innovations. Extrapolating our measured scaling exponents to 2023 compute budgets (\num{$1.3 \times 10^{22}$} FLOPs), the LSTM to Kaplan Transformer transition alone accounts for \num{$\lstmKaplanCEG$} of the efficiency gains. Combined with the Chinchilla rebalancing gains of \num{$\chinchillaCEG$} and \num{$\scaleInvCEG$} for scale-invariant innovations, we estimate total efficiency gains of approximately \num{$\totalCEG$}, helping close the gap with literature estimates over the same time period.

\paragraph{Algorithmic Versus Hardware Improvements} 

Algorithmic progress is traditionally seen as much faster than hardware progress. For instance, \citet{ho2024algorithmic} estimates algorithmic progress of $22,000$ over a ten year period while Moore's law over a ten year period would only amount to efficiency gains of $32\times$. However, at small compute scales, for instance using $10^{16}$ FLOPs, algorithmic progress could be only \num{$20\times$} -- less than hardware progress (See Figure~\ref{fig:stackplot}).

\paragraph{The Concentration of Progress in Architectural Transitions}

Our decomposition reveals that algorithmic progress exhibits extreme concentration: the LSTM-to-Transformer architectural shift comprises $\num{\transitionShare}$ of measured log-efficiency gains at 2023 compute levels, with the Kaplan-to-Chinchilla rebalancing accounting for most of the remainder. The distribution of scale-invariant innovations is highly skewed, with a handful of improvements approaching or exceeding $2\times$, while most innovations contribute less than $1.5\times$. 

This concentration has two important implications. First, algorithmic progress appears more punctuated than smooth exponential trends would suggest, with long periods of incremental refinement interrupted by rare architectural transitions. Second, the small magnitude of most scale-invariant improvements ($<10\times$ cumulatively) suggests that future progress may similarly depend on discovering fundamentally new architectures rather than incremental refinements of existing ones.

\subsection{Implications for the Future of AI}

Many forecasts of AI capabilities assume smooth exponential gains in training efficiency over time \citep{AI2027,erdil2025gate} regardless of compute level. Our results suggest critical corrections to this view.

\paragraph{Larger Players Gain More From Algorithms} 
If efficiency improvements are scale-dependent, then algorithmic progress has not been uniform across small and large algorithmic providers. Model trainers with a large computational budget see greater improvements with the introduction of new algorithms than smaller builders. Model builders scaling along the compute frontier see gains in the tens of thousands, while smaller model builders see algorithmic progress at levels much more similar to hardware progress.

\paragraph{Reference-dependence of Progress Measurements.} 
In settings with scale-dependent algorithms, the apparent rate of “algorithmic progress” is not intrinsic—it depends on the reference baseline. The same sequence of frontier MoE Transformers can show rapid CEG-multiplier growth when measured against LSTMs (e.g., $\sim 63\%$/year) yet appear flat when measured against dense Transformers. Thus, meaningful efficiency trends require specifying a fixed reference algorithm; time-series estimates anchored to older baselines may overstate progress relative to contemporary models, even at the frontier.

\paragraph{Difficulty of Early Identification} Scale-dependent innovations may appear modest at small scales, making it difficult to identify transformative improvements before committing substantial compute resources.  We estimate substantially larger gains from LSTM to Transformer and Kaplan to Chinchilla scaling at higher scales. This suggests that limits to compute scaling pose obstacles not only to realizing efficiency gains but also to discovering them. This opens the possibility of algorithms that decrease performance at small scales, but which would improve performance if we scaled. Therefore, researchers must examine the scale-dependent nature of their improvements to identify the real impact of their innovations.

\paragraph{Will Algorithmic Progress Slow Down?} These findings have important implications for society and AI governance. If efficiency gains primarily arise from scale-dependent innovations, then limits to compute scaling---whether from energy constraints, semiconductor supply chains, or regulatory restrictions---may substantially slow AI algorithmic progress. In addition, if algorithmic progress cannot be separated from compute investment, then the current exponential growth in AI capabilities may be more brittle than commonly assumed, requiring sustained increases in both computational resources and algorithmic breakthroughs.

\begin{ack}
\paragraph{Author Contributions.}
Hans led the project, including designing and running the experiments, setting research direction, developing conclusions and analysis, and taking primary responsibility for writing and submitting the project. Alex originated and wrote the sections on reference dependence in algorithmic progress and the mathematical generalization of CEG multipliers. He also contributed to figures, numerical analysis (particularly Appendix~\ref{appendix:kaplan_chinchilla}), writing, and editing. Jayson, Ana, Jonathan, and Neil contributed advising, figure design, and editing. Neil supervised the project. Anmol collected algorithmic innovations included in the study. 

\paragraph{Acknowledgments.}
We thank Zachary Brown and Yonatan Belinkov for insightful comments on this manuscript. We also thank William Moses for advising the early stages of this project. 

\paragraph{Funding.}
This work is funded by Open Philanthropy/Good Ventures. A.T. and A.F. are funded by the Alfred P. Sloan Foundation (G-2025-25164) and Microsoft. 
The authors acknowledge the MIT SuperCloud and Lincoln Laboratory Supercomputing Center for providing HPC resources that have contributed to the research results reported within this paper.

\paragraph{Competing interests.}
The authors declare no competing interests.

\end{ack}

\clearpage
\bibliographystyle{icml2026} 
\bibliography{references}

\appendix
\section{Model Architecture}\label{appendix:model_architecture}

\begin{table}[hb]
\centering
\caption{Base hyperparameters used to train the Modern Transformer model. Taken from \citet{porian2024resolving} and optimal recommendations from \citet{cs336_spring2025_lecture3}. }
\label{tab:transformer_hyperparams_modern} 
\vspace{.1cm}
\small
\begin{tabularx}{\columnwidth}{@{}lX@{}}
\toprule
\textbf{Hyperparameter} & \textbf{Value} \\
\midrule
Width/Depth aspect ratio      & 16 \\
Norm type                     & RMSNorm prenorm \\
Feedforward / model dimension & 4 \\
Activation function           & SwiGLU \\
Positional encoding           & Rotary \\
Sequence length               & 128 \\
Stride                        & 128 \\
Batch size                    & 128\footnotemark \\
Dropout                       & 0.0 \\
Weight decay                  & 0.01 \\
Gradient clipping             & 1.0 \\
Learning rate schedule        & cosine annealing \\
Min learning rate             & 0.1$\times$ max lr \\
Warmup fraction               & $10\%$ \\
Default optimizer             & AdamW \\
$\beta_1$                     & 0.9 \\
$\beta_2$                     & 0.999 \\
Initialization                & GPT-2/BERT init \\
Embedding weight tying        & True \\
Tokenizer                     & GPT-2 Tokenizer \\
Vocabulary                    & 50257 \\
Dataset                       & C4 \\
Validation set size           & 500k tokens \\
Token-to-parameter ratio      & 40 \\
\bottomrule
\end{tabularx}
\footnotetext{Ablation experiments used a batch size of 64 as these were all done at a small scale.}
\normalsize
\end{table}

\begin{table}[tb]
\centering
\caption{Architecture and hyperparameters for our Retro Transformer. This matches Table~\ref{tab:transformer_hyperparams_modern} with key algorithmic choices reverted.}
\label{tab:transformer_hyperparams_retro}
\vspace{.1cm}
\small
\begin{tabularx}{\columnwidth}{@{}lX@{}}
\toprule
\textbf{Hyperparameter} & \textbf{Value} \\
\midrule
Width/Depth aspect ratio      & 16 \\
Norm type                     & layer postnorm \\
Feedforward / model dimension & 4 \\
Activation function           & GeLU \\
Positional encoding           & Sinusoidal \\
Sequence length               & 128 \\
Stride                        & 128 \\
Batch size                    & 128 \\
Dropout                       & 0.0 \\
Weight decay                  & 0.01 \\
Gradient clipping             & 1.0 \\
Learning rate schedule        & inverse square root decay \\
Min learning rate             & 0.1$\times$ max lr \\
Warmup fraction               & $10\%$ \\
Default optimizer             & AdamW \\
$\beta_1$                     & 0.9 \\
$\beta_2$                     & 0.999 \\
Initialization                & GPT-2/BERT init \\
Embedding weight tying        & True \\
Tokenizer                     & GPT-2 Tokenizer \\
Vocabulary                    & 50257 \\
Dataset                       & C4 \\
Validation set size           & 500k tokens \\
Token-to-parameter ratio      & 40 \\
\bottomrule
\end{tabularx}
\normalsize
\end{table}

\begin{table}[tb]
\centering
\caption{Base hyperparameters used to train the LSTM model, mostly taken from \citet{melis2017state}. This includes TBPTT length, weight decay, number of layers, batch size, etc. \citet{melis2017state} recommends setting $\beta_1$ to 0, we find dropout is not helpful, see Appendix~\ref{sec:hyperparameter_discussion}}
\label{tab:lstm_hyperparams}
\vspace{.1cm}
\small
\begin{tabularx}{\columnwidth}{@{}lX@{}}
\toprule
\textbf{Hyperparameter} & \textbf{Value} \\
\midrule
Number of layers            & 1 \\
TBPTT length                & 32 \\
TBPTT stride                & 32 \\
Sequence length             & 128 \\
Batch size                  & 128 \\
Input dropout               & 0.0 \\
Hidden dropout              & 0.0 \\
Output dropout              & 0.0 \\
Weight decay                & $10^{-4}$ \\
Gradient clipping           & 1.0 \\
Learning rate schedule      & cosine annealing \\
Min learning rate           & 0.1$\times$ max lr \\
Warmup fraction             & $10\%$ \\
Default optimizer           & AdamW \\
$\beta_1$                   & 0.0 \\
$\beta_2$                   & 0.999 \\
Initialization              & LSTM (orthogonal hidden matrices) \\
Embedding weight tying      & True \\
Tokenizer                   & GPT-2 Tokenizer \\
Vocabulary                  & 50257 \\
Dataset                     & C4 \\
Validation set size         & 500k tokens \\
Token-to-parameter ratio    & 40 \\
\bottomrule
\end{tabularx}
\normalsize
\end{table}

 \subsection{Learning Rate Selection}\label{sec:learning_rate_tune}                                                                                                              
  For each architecture (modern transformer, historical transformer, LSTM 1-layer, LSTM 2-layer), we perform a dedicated learning rate sweep to identify how the optimal learning rate $\eta^*$ scales with model size. At each of 11 hidden dimensions $d   
  \in \{32, 48, 64, \ldots, 256\}$, we train with 8 candidate learning rates spaced 0.25 decades apart, using 5\% of the Chinchilla-optimal token budget per trial. The optimal learning rate per dimension is estimated via a parabolic vertex method
   \citep{bjorck2025scaling, vaultgemma2025}: we fit a quadratic $L = a(\ln \eta)^2 + b(\ln \eta) + c$ to the final validation losses across the swept grid and take the analytic vertex as $\eta^*$, falling back to the grid argmin when the parabola is          
  non-convex or the vertex lies outside the swept range.                                                                                                                                                                                                     
                                                         
  We then regress $\log_{10} \eta^*$ against $\log_{10} N$, where $N$ is the non-embedding parameter count, obtaining a power-law scaling law $\eta^*(N) = A \cdot N^{\alpha}$ separately for each architecture. This fit is used to
  extrapolate a learning rate for every model size used in the scaling experiments, providing a principled, architecture-specific alternative to per-run hyperparameter tuning.

\subsection{Further Discussion of Hyperparameter Choices Including Dropout}\label{sec:hyperparameter_discussion}

We choose to use no dropout for both our LSTM and Transformer as both perform considerably better without dropout. To check this for LSTM we compared an LSTM with hidden dimension 64 with the dropout settings recommended by \citet{melis2017state} to an LSTM with no dropout and tune learning rates for both. The best performing LSTM with dropout has a validation loss of 6.5, while the best performing LSTM without dropout has a validation loss of 5.9.

\subsection{Data Selection, Loading, and Size}
We choose to use a subsample of the C4 dataset for both the transformer and LSTM.  The size of dataset we use for training is based on the model size and our token-to-parameter ratio. In order to better outline the optimal compute frontier, we choose to use a token-to-parameter ratio of 40 in our scaling experiments. This is to ensure that we do not stop early before compute-optimal level of data. Stopping early would change the frontier envelope, while stopping late would not change the frontier, as overtrained models would pass the frontier. While for our ablation experiments, we use a token to parameter ratio of 20 as recommended by \citet{hoffmann2022trainingcomputeoptimallargelanguage}.

\section{Scaling-Experiment Models and Hyperparameters}
\label{app:scaling_models_details}

This appendix records the model definitions and hyperparameters used in Section~\ref{sec:experimental_setup}. Unless otherwise stated, all models are trained on the same dataset and tokenizer, with matched optimization and tuning budgets across model families.
\FloatBarrier
\section{2-Layer LSTM Baseline}
\label{app:2_layer_baseline}
To check that our results do not arise from incorrectly choosing the wrong layer number baseline. We run a 2-layer LSTM baseline scaling using the same setup as our LSTM model with a batch size of 128. Overall, we found that 2-layers performs worse than a 1-layer LSTM for all scales. This is supported by literature, which generally finds better performance of 1-layer LSTMs at small scales (\citep{melis2017state}, \citep{droppo2021scaling}).

\begin{figure}[h!]
    \centering
    \includegraphics[width=0.5\linewidth]{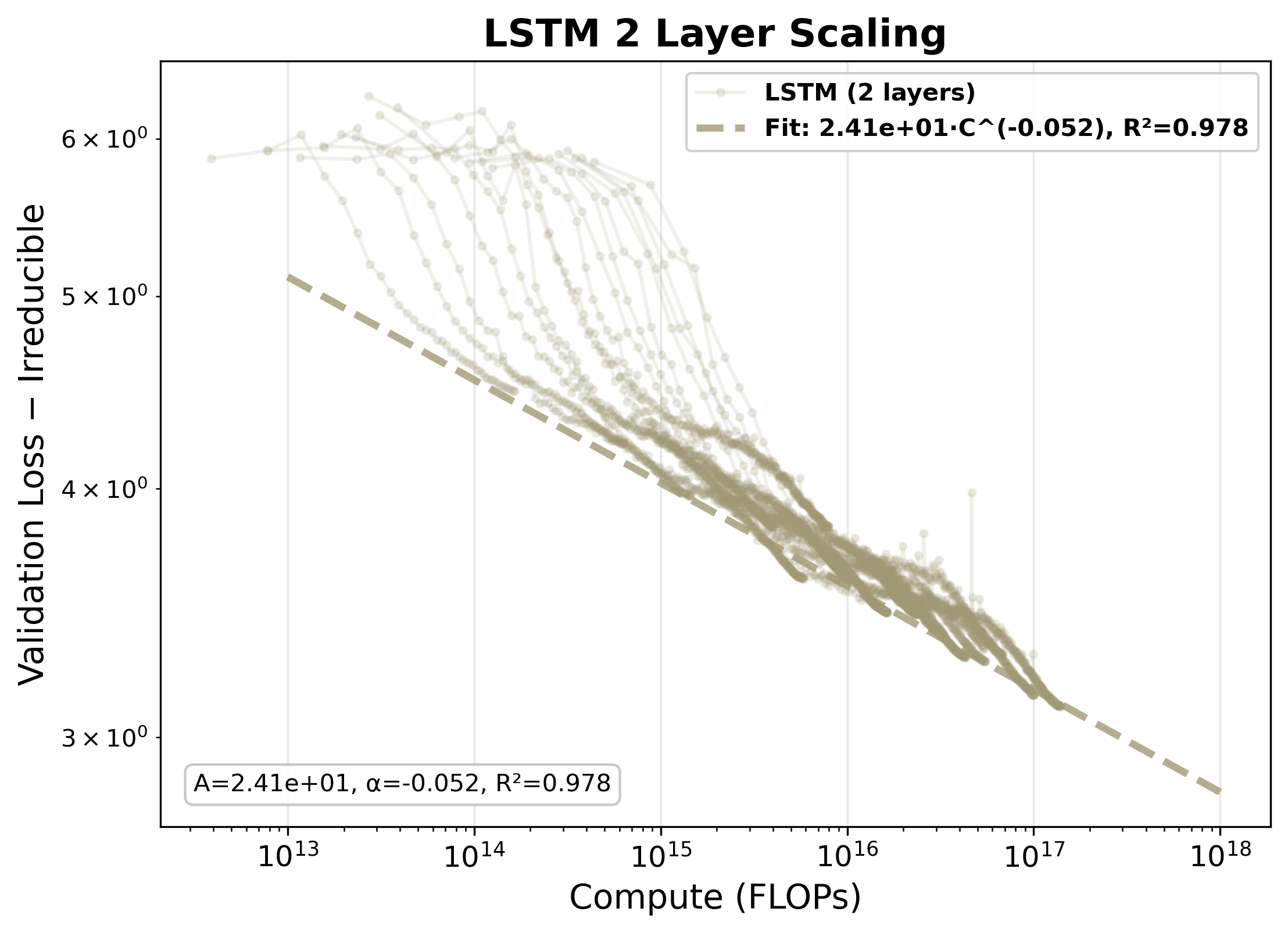}
    \caption{2-layer LSTM has lower scaling exponent than our 1-layer LSTM baseline. See Appendix~\ref{app:2_layer_baseline}.}
    \label{fig:layer2baseline}
\end{figure}

\subsection{Model Definitions}\label{appendix:model_definitions}

\paragraph{Modern Transformer.}
Decoder-only Transformer with contemporary default components: rotary positional encodings, RMSNorm prenorm, and SwiGLU. We keep a constant width-to-depth aspect ratio across scales (chosen for stability in the small-model regime), and set the attention-head configuration to maintain a small head dimension appropriate for these sizes. Weight initialization follows the standard GPT-2/BERT-style recipe (zero biases; Gaussian weights). FLOPs are measured using a PyTorch profiler on one minibatch and scaled by the number of optimizer steps and gradient-accumulation schedule.

\paragraph{Retro Transformer.}
Matches the Modern Transformer architecture and scaling rules but reverts four components to approximate an early Transformer-style recipe:
(i) SwiGLU $\rightarrow$ GeLU,
(ii) RMSNorm prenorm $\rightarrow$ LayerNorm postnorm,
(iii) rotary $\rightarrow$ sinusoidal positional encodings,
(iv) cosine decay $\rightarrow$ inverse square-root decay.

\paragraph{Kaplan Transformer (analytical scaling policy).}
The Kaplan Transformer is \emph{not} a separate architecture. It is the Retro Transformer evaluated under Kaplan-style parameter--data allocation via an analytical adjustment described in Appendix~\ref{appendix:kaplan_chinchilla}. We use this to isolate the effect of scaling policy from architectural changes.

\paragraph{LSTM baseline.}
Vanilla LSTM based on the strong pre-Transformer recipe of \citet{melis2017state}. We use one LSTM layer throughout scaling for the reasons discussed in Section~\ref{sec:experimental_setup}. We use Xavier-uniform embeddings, orthogonal recurrent matrices, and a forget-gate bias of $+1$. Following \citet{melis2017state,merity2017regularizing}, the original recipe uses variational dropout; in our internet-scale regime we set dropout to zero (as in common Transformer practice \citep{hoffmann2022trainingcomputeoptimallargelanguage}) and tune learning rates comparably across families.

\subsection{Scale Grid and Parameter Ranges}
We scale models by varying hidden size on a fixed grid and holding aspect ratios constant within each family. In Section~\ref{sec:scale_dependent} we use hidden dimensions
\[
[16, 32, 48, 64, 80, 96, 112, 128, 160, 192, 224, 256],
\]
corresponding to Transformer sizes of approximately 1.6M--29M parameters. For our LSTM model we choose a slightly larger range of hidden dimension sizes so that we can achieve a similar FLOP range as our transformer model: 
\[
[16, 32, 48, 64, 80, 96, 112, 128, 160, 192, 224, 256,
    288, 320, 352, 384]
\]
This corresponds to LSTMS of around 1.6M--20.5M parameters (exact counts depend on embedding and vocabulary parameters).

\subsection{Optimization, Data, and Tuning Protocol}
All models use the same tokenizer (GPT2 Tokenizer) and dataset (C4) and the same validation protocol. We run matched learning-rate tuning across model families: we evaluate a coarse learning-rate grid at multiple sizes, then fit a smooth learning-rate vs.\ hidden-size relationship used to set learning rates at intermediate sizes. Full tuning details are in Appendix~\ref{sec:learning_rate_tune}. Where relevant, we report compute using measured FLOPs rather than step counts.

\section{Reproducibility, Code, and Compute Statement}
All code used in model training and model analysis, along with setup instructions, is hosted here:
\url{https://github.com/hansgundlach/Experimental_Progress}. The main body of our text contains details on our training and analysis procedure. A detailed list of hyperparameters used is in Appendix~\ref{appendix:model_architecture}. All code and data are under CC BY 4.0 license.
Final experiments were done over three days on 8 V100 GPUs. Each full scaling run took roughly 12 hours.

\FloatBarrier

\section{Sensitivity Analysis}\label{sec:sensitivity_analysis}
\begin{figure*}[t]
    \centering
    \includegraphics[width=\textwidth]{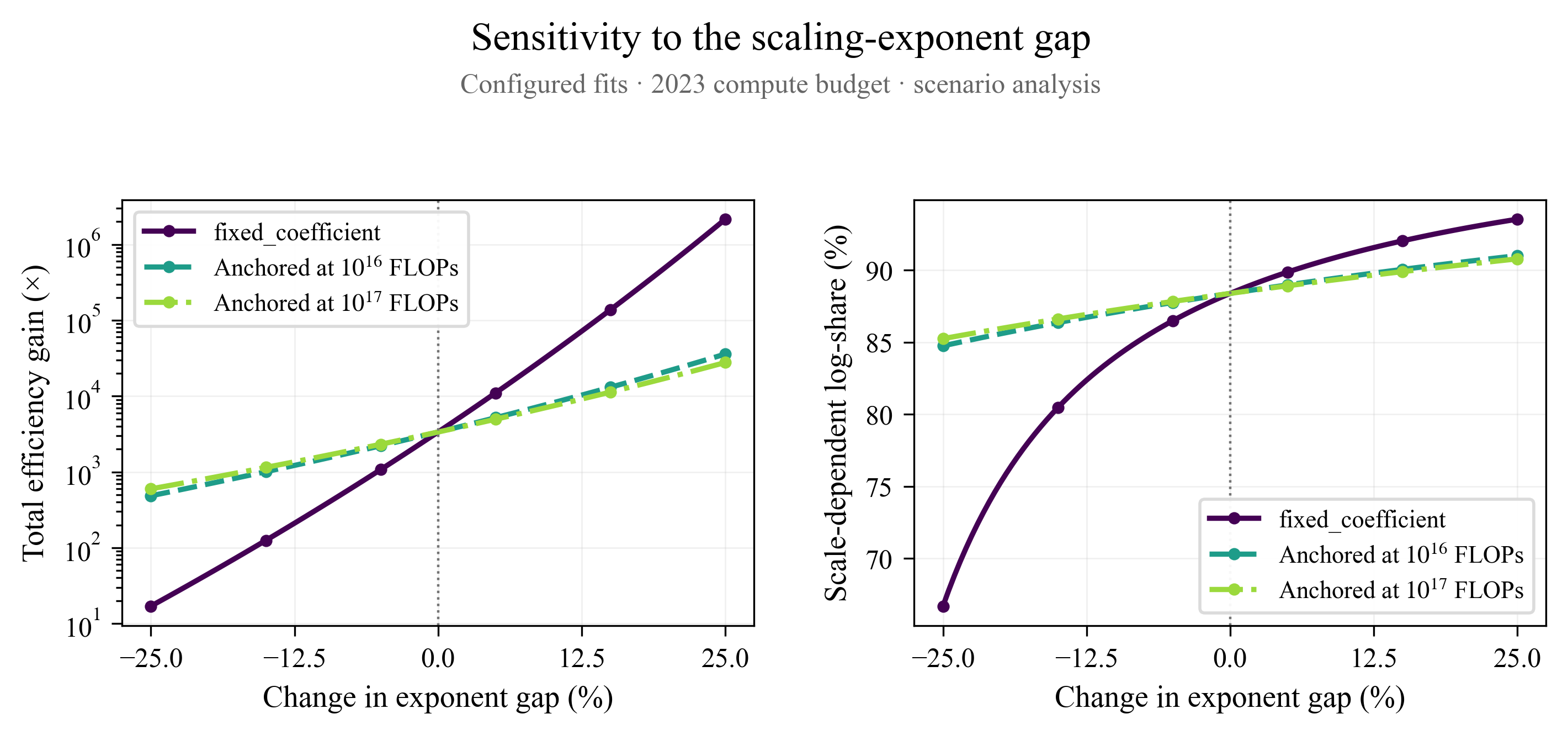}
    \caption{Sensitivity of the extrapolated 2023 total efficiency gain (left) and the scale-dependent log-share (right) to a $\pm 25\%$ change in the LSTM--Transformer exponent gap. Holding the power-law coefficient fixed moves both quantities much more than anchoring the perturbed curve at $10^{16}$ or $10^{17}$ FLOPs.}
    \label{fig:exponent_gap}
\end{figure*}

A key parameter in our model of algorithmic progress is the scaling-exponent difference between Transformers ($\alpha_{TM}$) and LSTMs ($\alpha_L$). We vary this difference and measure the effect on our overall extrapolated efficiency number and the scale dependent share of algorithmic progress. The sensitivity analysis depends on how we anchor the extrapolation. Perturbing the scaling exponent while holding the coefficient fixed also changes predicted losses throughout the measured compute range, conflating slope sensitivity with changes in baseline performance. Anchoring the perturbed curve at a compute level within the experimental range preserves the experimentally measured loss at that value, allowing us to isolate how alternative slopes affect extrapolated efficiency gains as the exponent changes. If we anchor at $10^{16}$ FLOPs, by taking the experimental loss at that value as given and change the exponent used at extrapolation, the total efficiency gap ranges from 485x to 35,860x (as opposed to \num{$\totalCEG$}). Varying the exponent gap without changing any other coefficient (equivalent to anchoring at 1 FLOP) has a large effect on our overall efficiency extrapolation number, leading to total efficiency ranging from 17x to 2.16 million. Regardless, of the technique used we see that the LSTM-to-Transformer transition and Kaplan to Chinchilla account for the majority of measured progress (log or non-log). The impact of each of these perturbations on absolute and log efficiency gains is shown in Figure~\ref{fig:exponent_gap}.

\begin{figure*}[t]
    \centering
    \includegraphics[width=\textwidth]{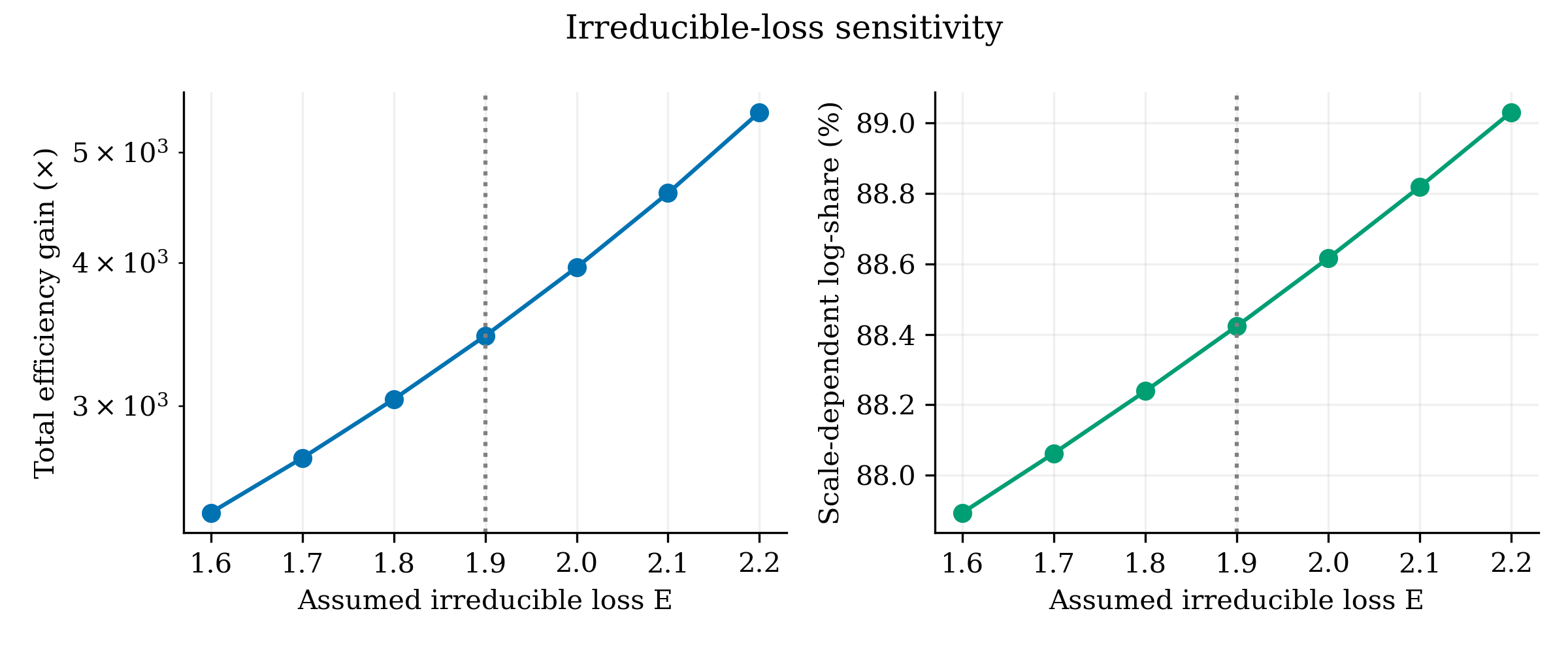}
    \caption{Refits of the total efficiency gain (left) and the scale-dependent log-share (right) under assumed irreducible losses $E$ from 1.6 to 2.2. The dashed line marks the reference value $E = 1.9$ used in the main fits.}
    \label{fig:irreducible_loss_refits}
\end{figure*}

\begin{figure}[t]
    \centering
    \includegraphics[width=0.65\linewidth]{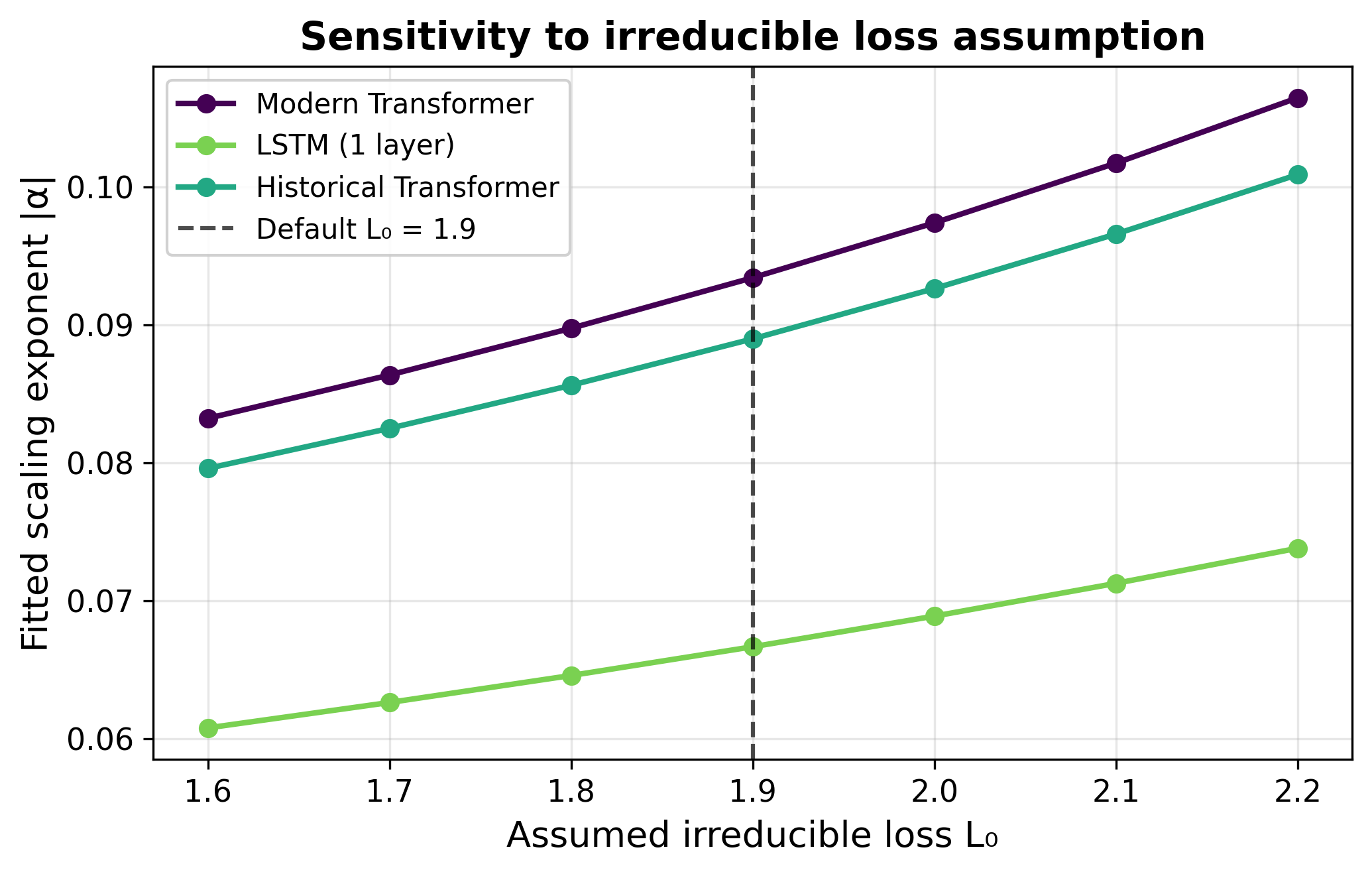}
    \caption{Sensitivity of fitted scaling exponents to the assumed irreducible loss.}
    \label{fig:irreducible_loss_sensitivity}
\end{figure}

We use a fixed Irreducible loss of 1.9 for our scaling law fits; this is fit and analyzed in Appendix~\ref{sec:irreducible_loss}. However, within a reasonable range, our choice of irreducible loss has relatively little effect on the scaling exponents we determine (see Figure~\ref{fig:irreducible_loss_sensitivity}). It is also much less important to the total efficiency gains we measure (see Figure~\ref{fig:irreducible_loss_refits}). 

\FloatBarrier
\subsection{Fit Quality}\label{app:fit_statistics}

Table~\ref{tab:fit_statistics} reports the fitted exponents and goodness-of-fit statistics for the power-law fits used in Section~\ref{sec:scale_dependent}. The fits are highly accurate ($R^{2} \geq 0.98$) with tight 95\% confidence intervals on the fitted exponents (at most $\pm 0.001$). Consequently, we expect the dominant sources of uncertainty in our extrapolations to come from hyperparameter tuning, the choice of irreducible loss, and dataset effects (Appendix~\ref{sec:irreducible_loss} and Section~\ref{sec:data}) rather than from the fitting procedure itself.

\begin{table}[h]
\centering
\begin{tabular}{lccc}
\toprule
Architecture & Exponent $\alpha$ (95\% CI) & Coefficient $A$ & $R^2$ \\
\midrule
Modern Transformer & $-0.093 \pm 0.000$ & $86.2$ & $0.999$ \\
LSTM (1 layer) & $-0.067 \pm 0.001$ & $39.7$ & $0.980$ \\
Historical Transformer & $-0.089 \pm 0.000$ & $74.9$ & $0.998$ \\
\bottomrule
\end{tabular}
\caption{Power-law fit statistics for the scaling experiments in Section~\ref{sec:scale_dependent}. We fit $L - L_\text{irr} = A \cdot C^{\alpha}$ to the compute-optimal frontier of each model family.}
\label{tab:fit_statistics}
\end{table}

\FloatBarrier

\section{Kaplan-Style Scaling When the True Frontier is Chinchilla-Optimal}\label{appendix:kaplan_chinchilla}

In Figure \ref{fig:kaplan-v-chinch} below, we show loss curves from using Kaplan and Chinchilla's recommendations for compute optimal scaling (specifically optimal scaling by analytically solving the full parameter model). We take the Chinchilla functional form for model loss $L(N,D)$ as ground-truth, only modifying the allocation of parameters and dataset size between each curve. Unsurprisingly, Chinchilla scaling is a perfect power law.

We use the constants $E = 1.69$, $A = 406.4$, $B = 410.7$, $\alpha = .34$, and $\beta = .28$ and include the functional form below for the reader's convenience.
\begin{align}
L_\text{Chinchilla}(N, D) &= \frac{A}{N^{\alpha}} + \frac{B}{D^{\beta}} + E \\
&= \frac{406.4}{N^{0.34}} + \frac{410.7}{D^{0.28}} + 1.69
\end{align}

For the compute allocation, Chinchilla scaling explicitly defines optimal scaling of $N$ and $D$ to be

\begin{align}
N_{\text{Chinchilla}}(C) &=  G\left(\frac{C}{6}\right)^{\beta/(\alpha + \beta)}\\
D_{\text{Chinchilla}}(C) &= \left( \frac{1}{G} \right)\left(\frac{C}{6}\right)^{\alpha/(\alpha + \beta)}
\end{align}

where $G = (\frac{\alpha A}{\beta B})^{1/(\alpha + \beta)}$. By construction, this gives a precise power law with a well defined inverse:
\begin{equation}
\text{ChinchillaCompute}(L) = \frac{5.4 \times 10^{19}}{(L - 1.69)^{6.5}}
\end{equation}
Finally, we infer the full Kaplan scaling recommendation: Figure 14 in \citet{kaplan2020scalinglawsneurallanguage} gives a functional form for compute optimal dataset scaling, and after applying unit conversions and using the relationship $C = 6ND$, we get the following:

\begin{align}
N_{\text{Kaplan}}(C) &=  (3.6 \times 10^{-6})C^{0.73}\\
D_{\text{Kaplan}}(C) &= (4.6 \times 10^4)C^{0.27}
\end{align}

Together, these give us a closed form for the compute multiplier when rebalancing from Kaplan to Chinchilla scaling for a fixed $C$. We call this function $M$, and compute it as follows:

\begin{align}
M(C) &= \frac{C}{\text{ChinchillaCompute}(L_\text{Chinchilla}(N, D))} \nonumber\\
&\quad \text{where } N = N_{\text{Kaplan}}(C),\ D = D_{\text{Kaplan}}(C) \nonumber\\
&= (1.85 \times 10^{-20})(C)\left(\frac{28737.9}{C^{0.2482}} + \frac{20.337}{C^{0.0756}}\right)^{6.512}
\end{align}

As $C$ gets large (roughly $10^{23}$ FLOPs), we can use a power law to approximate this expression:
$$M(C) = (6.13 \times 10^{-12})(C^{0.508})$$

 \begin{figure}[h!]
     \centering
     \includegraphics[width=\linewidth]{ 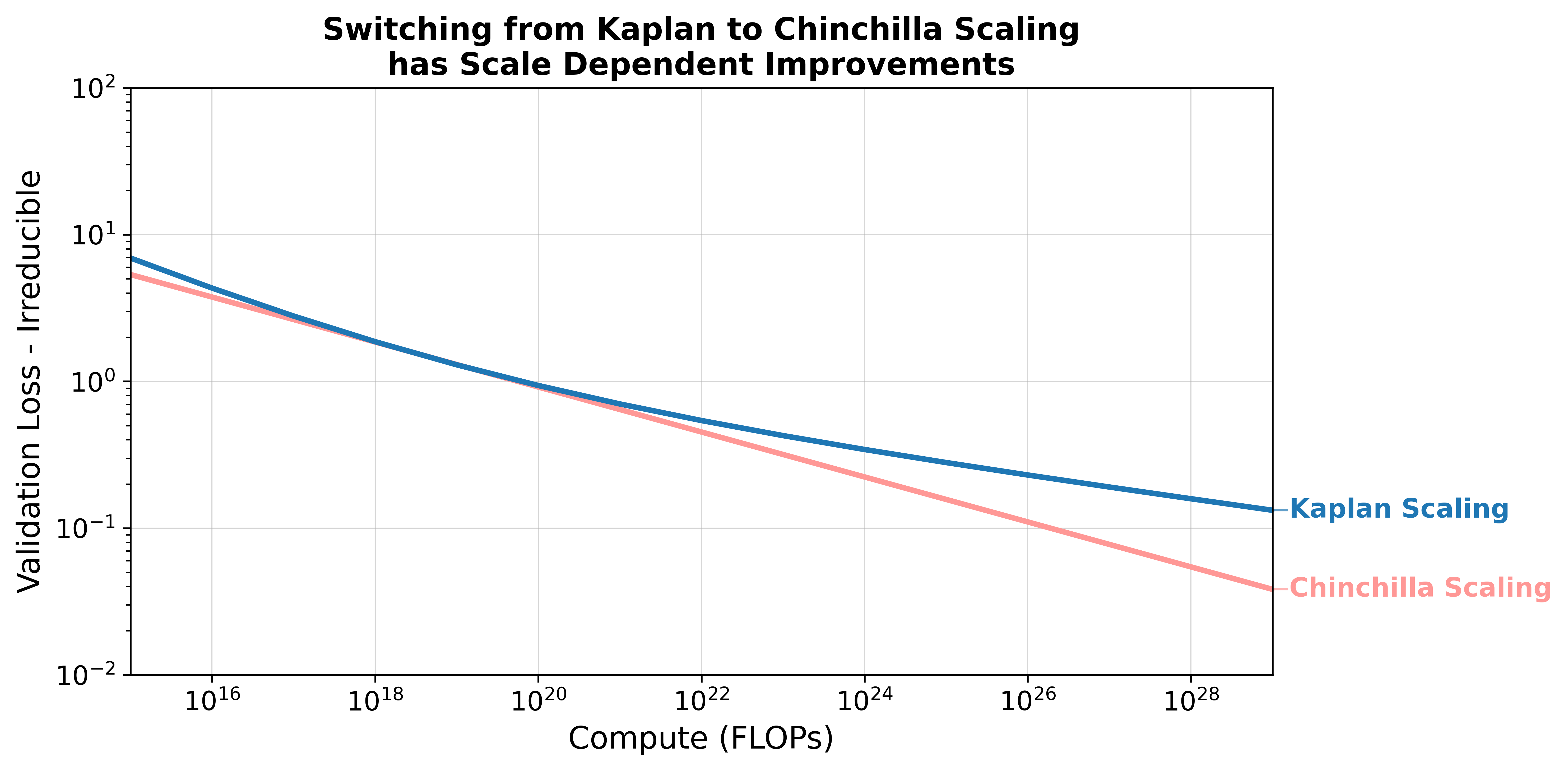}
     \caption{Analytical difference in performance between transformer models scaled with Kaplan versus Chinchilla recommendations. Interestingly, the efficiency gap first converges, then diverges.}
     \label{fig:kaplan-v-chinch}
 \end{figure}
\FloatBarrier
\section{Transformer SGD Scaling Study}
\label{sec:sgd_scaling}

In order to train a Transformer with SGD we use a momentum of .98 and 0 weight decay as recommended by \citet{sreckovic2025your}. We use the same scaling procedure as our default Transformer, and all other hyperparameters are held constant. We find little scaling difference between Transformers trained with SGD and Transformers trained with AdamW.  However, in our experiments, transformers trained with SGD are notably more unstable. For instance, our SGD training curve has a slight concavity early on.

\begin{figure}[h!]
    \centering
    \includegraphics[width=\linewidth]{ 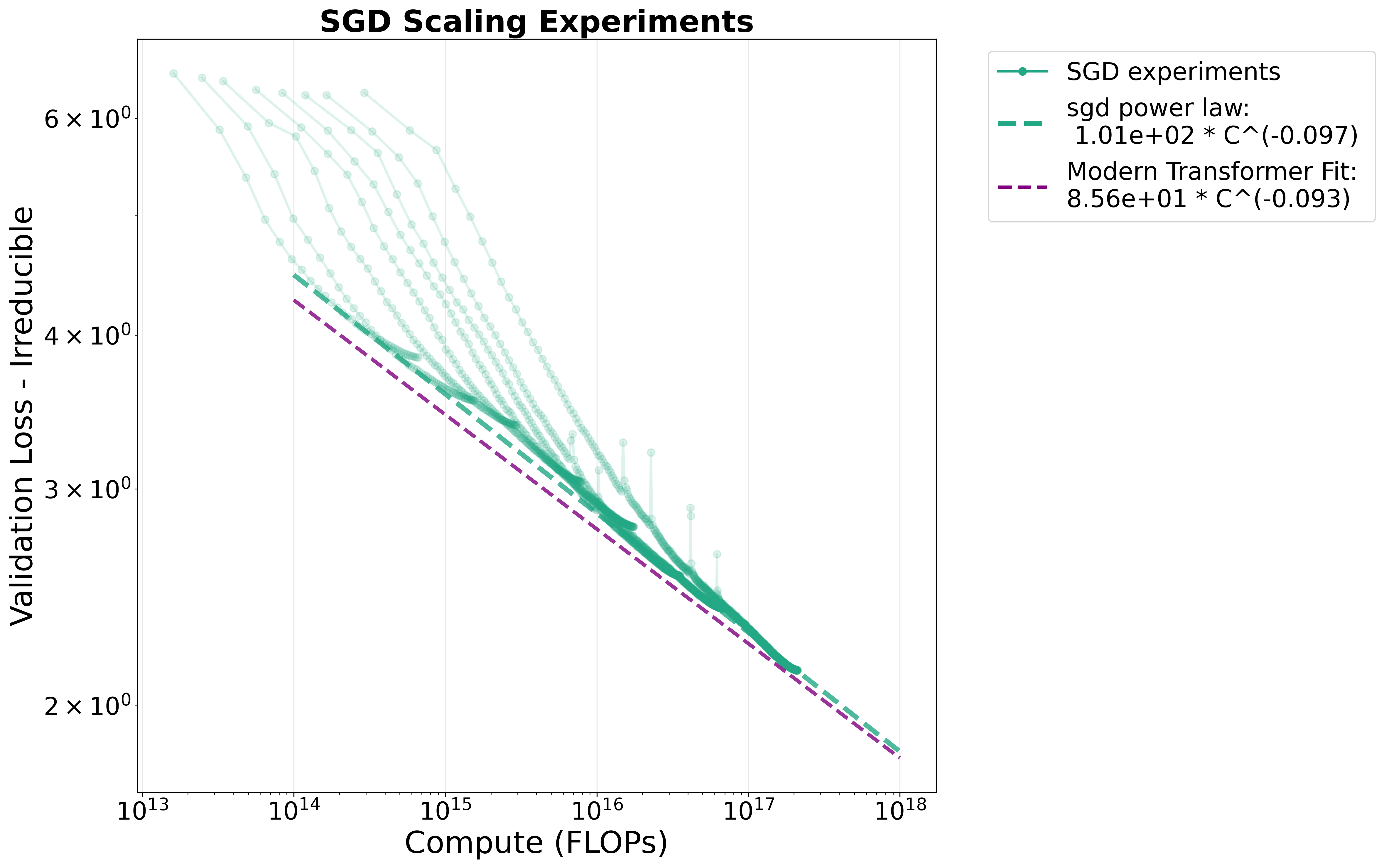}
    \caption{Optimizers have little effect on scaling exponents. The blue dashed line represents the compute-optimal scaling fit for our Modern Transformer trained with SGD. The purple dashed line represents the experimental fit for our Transformer trained with AdamW.}
    \label{fig:placeholder}
\end{figure}

\FloatBarrier
\section{Irreducible Loss}\label{sec:irreducible_loss}
The irreducible loss represents the self-entropy of the data distribution. This means it varies from dataset to dataset but should be constant across models.  In our paper, we set the irreducible loss to 1.9.  In contrast, \citet{hoffmann2022trainingcomputeoptimallargelanguage} fits an irreducible loss of 1.69 using MassiveText (similar to C4) \citet{besiroglu2024chinchilla} finds an irreducible loss of 1.8 using \citet{hoffmann2022trainingcomputeoptimallargelanguage} data. \citet{porian2024resolving} did a replication of Kaplan and Chinchilla scaling on RefinedWeb and OpenWebText2. They find irreducible loss fits from 1.68 to 2.01. \citet{muennighoff2023scaling} finds an irreducible loss of 1.87 for Transformers trained on C4. In light of this variation, we use scikit learn to fit the optimal frontier to the form $L = E+AC^{-\alpha}$, restricting $E$ to lie between 1.3 and 2.2. For our Modern Transformer, 2017 transformer, and LSTM, this approach yields an irreducible loss close to 1.9 for all the models used in our study. We also do a robustness check of the irreducible loss choice in Appendix~\ref{sec:sensitivity_analysis}.

\section{Reference-Dependence of CEG Multipliers}\label{app:reference_dependence}

Figure~\ref{fig:reference_models} visualizes the reference-dependence argument in Section~\ref{sec:implications}. We compare the apparent CEG multiplier and annual growth rate for the same sequence of MoE-enhanced Modern Transformers under three choices of reference algorithm: LSTMs, Kaplan Transformers, and dense Transformers. The resulting trend differs sharply across reference points even though the underlying sequence of models is unchanged.

\section{Formal Definition of CEG Functions and Multipliers}\label{app:ceg_formal}

In order to more precisely discuss algorithmic progress, we introduce a generalized notion of the standard compute equivalent gains (CEG) framework \citep{davidson2023aicapabilitiessignificantlyimproved,ho2024algorithmic,sanderson2025rethinking}. Our approach varies from much of the existing literature by allowing algorithmic innovations to have different efficiency gains across compute scales. More precisely, we define the \textbf{CEG function} and \textbf{CEG multiplier} distinctly:

\begin{center}
\textit{Definition (CEG function).} For two algorithms $A$, $A'$ and a fixed performance metric (e.g.\ loss), the \textbf{compute equivalent gain function} from $A$ to $A'$ is a function $f$ such that a model trained with $A$ using compute $C$ and a model trained with $A'$ using compute $C / f(C)$ reach equivalent performance, for all $C$ in the domain of $f$.
\end{center}

Fixing the compute budget for one of the algorithms, we can recover the traditional definition of a CEG multiplier:

\begin{center}
\textit{Definition (CEG multiplier).} Let $M$ be a model trained with algorithm $A$ using compute $C$. Let $f$ be the CEG function between $A$ and some $A'$. Then $f(C)$ is the \textbf{CEG multiplier} of $A'$ \textit{relative to $M$}.
\end{center}

Intuitively, CEG functions compare training algorithms across compute scales, whereas CEG multipliers compare models by fixing a reference compute (or performance) level.\footnote{``Training algorithms'' includes everything except compute, e.g.\ architecture, optimizers, and data--parameter balancing.} This distinction is central to our analysis: when algorithms exhibit scale dependence, reporting only CEG multipliers can induce misleading trends.\footnote{In our ablation experiments, we report CEG multipliers between algorithms using compute-optimal performance within our hyperparameter search.}

\paragraph{Edge cases.}
If an algorithm plateaus in performance, the CEG function may be undefined at some thresholds (informally requiring ``infinite'' compute) \citep{erdil2022algorithmic}. Finally, if the CEG function is known to be constant between two algorithms, one may choose any common performance threshold to estimate that constant value. For our ablation experiments, we argue the ablated innovations admit approximately constant CEG functions within our compute regime, and we use the lowest negative log-likelihood threshold achieved by all ablations (NLL$=5.3$) when reporting these multipliers.

\subsection{Interaction Between Algorithms}\label{sec:interaction_algs}

\begin{figure}[H]
    \centering
    \includegraphics[trim=0 -23.04bp 0 0,height=0.55\linewidth]{Figures/modern_transformer_improvements.png}\hfill
    \includegraphics[height=0.55\linewidth]{Figures/transformer_evolution_from_lstm.png}
    \caption{Enlarged copy of Figure~\ref{fig:interaction}. Left panel: comparison of the total effect of all measured post-Transformer algorithmic changes with the product of their individual ablation-estimated effects. Right panel: the same comparison for all measured changes, including the LSTM-to-Transformer transition. In each panel, the left bar shows the product of individual estimates and the right bar shows the measured combined effect. Combined gains are sub-multiplicative.}
    \label{fig:interaction_appendix}
\end{figure}

Conventional theory of algorithmic gains presupposes that most algorithmic gains are multiplicative \citep{ho2024algorithmic,davidson2023aicapabilitiessignificantlyimproved}. For instance, if algorithm $A$ increases training efficiency by $2\times$ and algorithm $B$ increases training efficiency by $3\times$, respectively, then the combination of algorithm $A$ and $B$ should have a training efficiency gain of $6\times$. However, across architectures this is not the case. For example, Adam provides a computational efficiency gain for transformers, yet most state-of-the-art LSTMs relied on RMSProp or plain SGD over Adam~\citep{merity2017regularizing,melis2017state,keskar2017improving}.

To help identify these interactions, we train a transformer model similar to the original architecture developed by \citet{vaswani2017attention}. In this ``Retro" Transformer, four algorithmic improvements are reverted. We switch from SwiGLU to GeLU, pre-RMSNorm to post-layernorm, rotary encoding to sinusoidal encodings, and cosine decay learning rate schedule to square root decay schedule. Using the estimates in Figure~\ref{fig:bar_improvements} and assuming multiplicative improvement we would expect CEG gains of \num{$3.55\times$}. However, the efficiency gain is only \num{$1.33\times$}. In addition, we compare our default or ``Modern" transformer (defined in Appendix~\ref{appendix:model_definitions}) to our LSTM baseline. At a loss threshold of $5.3$, we find a combined efficiency gain of  \num{$\smallScaleLSTMtoTM$}, whereas multiplying the individual improvements together, along with the LSTM to Retro Transformer improvements, yields a substantially higher predicted gain of \num{$\interactionPredictedCEG$}.\footnote{In the case where we have scaling fits, we use the CEG between scaling fits to be consistent with Figure~\ref{fig:scaling_graphs}. If we use individual training curves, we would get a Modern-to-Retro Transformer difference of \num{$2.7\times$} and a Modern-to-LSTM difference of \num{$5.9\times$}}

Figure~\ref{fig:interaction} summarizes these comparisons. This suggests that multiplying individual improvements, estimated from ablation experiments, overestimates the impact of algorithmic improvements.  We hypothesize the following explanation by analogy: consider replacing the engine of a modern car with a \textit{Model T} engine. A \textit{Model T} engine in an otherwise modern car might perform far worse than it did in the original \textit{Model T} due to system incompatibilities (similarly with any other component). As a result, one would overestimate the proportion of progress for each of these inventions independently.  Similarly, it may be that replacing algorithmic components individually appears to cause greater efficiency loss than replacing all components at once. In this sense, algorithmic improvements are sub-multiplicative and have slightly non-orthogonal effects.

\section{Ablation Experiment Details and Component-Level Results}
\label{app:ablation_details}

For our experimental approach, we examine a small 3.6M parameter Transformer with rotary-based encodings, a constant width-to-depth ratio, and a GeLU activation function. Our baseline is identical to our ``modern'' Transformer, described in our scaling experiments (Appendix~\ref{appendix:model_definitions}). For each ablation, we run learning-rate tunes by testing the default learning rate ($10^{-2.25}$) and $[10^{-2.5},10^{-2}]$.\footnote{If the optimal learning rate differs from the default, we test learning rates outside this range by factors of $\sqrt{10}$.} As mentioned previously, we choose the lowest loss threshold that all ablated models are capable of (in our case, a loss of 5.3) to determine the CEG multiplier. All of our ablation experiments use a token-to-parameter ratio of 20 as in \citet{hoffmann2022trainingcomputeoptimallargelanguage}.

\paragraph{Activation Functions.}
We test the impact of new activation functions like SwiGLU over the traditional GeLU activation function. In line with \citet{shazeer2020glu}, we find that SwiGLU has a measurable $1.17\times$ efficiency gain over GeLU even taking into account its increased parameter requirements.

\paragraph{Positional Encoding.}
We test three different types of positional encoding: rotary, sinusoidal, and learned positional encodings. We estimate that rotary encoding constitutes an improvement of $44\%$ in training efficiency over sinusoidal encoding originally employed in \citet{vaswani2017attention}. This is in contrast to \citet{sanderson2025rethinking}, which estimates a $70\%$ gain, and consistent with \citet{rope-eleutherai}, which sees a $10\%$--$30\%$ training efficiency gain. We expect this effect to be sequence-length dependent.

\paragraph{Learning-Rate Schedules.}
We test three prominent learning-rate schedules: linear warmup/inverse square-root decay \citep{raffel2020exploring,vaswani2017attention}, linear warmup/linear decay \citep{devlin2019bert}, and linear warmup with cosine decay \citep{brown2020language}. We observe very small training-efficiency differences between schedules ($<5\%$), consistent with \citet{kaplan2020scalinglawsneurallanguage}.

\paragraph{Normalization Techniques.}
We test a pre-RMSNorm model, a pre-layernorm model, and a post-RMSNorm model. Pre-layernorm(RMSNorm) applies layer normalization before attention in the residual stream and is generally seen as more stable than post-layernorm \citep{xiong2020layer}. RMSNorm removes mean-centering and can improve runtime \citep{zhang2019root}. We observe substantial improvements of $87\%$ transitioning from post-RMSNorm to pre-RMSNorm, with more modest improvements of $9\%$ switching from pre-layernorm to pre-RMSNorm.

\paragraph{Optimizers.}
Prior work has shown that SGD performs notably worse on Transformers compared to Adam \citep{ahn2023linear,zhao2024deconstructing}, potentially due to Hessian heterogeneity \citep{zhang2024transformers}. More recent results suggest the gap can narrow under specific hyperparameters \citep{sreckovic2025your}. With batch size 64, momentum 0.98, and no weight decay (as recommended by \citet{sreckovic2025your}), we observe a $1.67\times$ gap in training efficiency compared to AdamW. We further examine optimizer scaling in Appendix~\ref{sec:sgd_scaling}. We compare Adam to AdamW and find a negligible difference in our setting.

\section{Literature Estimates for Hard-to-Ablate Components}
\label{app:literature_estimates}

\paragraph{Why Include Literature Estimates?}
In addition to ablation experiments, we include conservative bounds from the literature for mixture-of-experts (MoE) training and tokenization methods. MoE models are challenging to implement and measure at small scales due to routing and loss-balancing, and tokenizers are difficult to assess using perplexity because perplexity is inherently tokenizer-dependent.

\paragraph{Mixture of Experts.}
\label{app:moe}
MoE is often discussed as an inference improvement, but it can also improve training efficiency. We approximate MoE training gains as $\le 2\times$ based on existing comparisons: \citet{riquelme2021scaling} report slightly less than $2\times$; isoflop comparisons in \citet{he2024mixture} suggest substantially less than $2\times$; \citet{li2025can} report approaching dense performance with roughly twice the compute; \citet{muennighoff2024olmoe} report larger gains but without controlled isoflop comparisons; and \citet{tay2023scaling} find similar scaling behavior with noisy results. Recently, researchers have started to find better ways to scale expert number for MoE systems. These improvements in scaling recipes might also be scale-dependent, but are outside the time range we study \citep{krajewski2024scaling, tillman2026scaledependent}. Given uncertainty, we leave a dedicated MoE scaling study to future work.

\paragraph{Tokenizers.}
Early language models used word- or character-level tokenization. Subword tokenization (BPE, WordPiece, SentencePiece) mitigated vocabulary/sequence-length tradeoffs. \citet{ali2024tokenizer} finds inefficient tokenization can increase training cost by $68\%$. We do not include tokenizer gains in our main aggregations due to uncertainty and limited standardized historical comparisons.

\end{document}